\documentclass[twoside,11pt]{article}

\usepackage{jmlr2e}

\newcommand\dotprod[2]{\langle #1, #2\rangle}  

\newcommand\R{{\mathbb R}}

\renewcommand\P{{\mathbb P}}        
\def\1{{\mathbf 1}}        

\newcommand\defeq{\coloneqq}

\abbreviation{GLM}{Generalized Linear Models}
\abbreviation{HMM}{Hidden Markov Model}
\abbreviation{HMC}{Hamiltonian Monte Carlo}
\abbreviation{AHMC}{Adaptive Hamiltonian Monte Carlo}
\abbreviation{MCMC}{Markov Chain Monte Carlo}
\terminology{GLMCTMC}{GLM-CTMC}
\abbreviation{GNR}{General Non-Reversible model}
\abbreviation{GTR}{General Time Reversible model}
\abbreviation{RMHMC}{Riemann Manifold Hamiltonian Monte Carlo}
\abbreviation{BPS}{Bouncy Particle Sampler}
\abbreviation{NUTS}{No U-Turn Sampler}
\abbreviation{ESJD}{Expected Squared Jumping Distance}
\abbreviation{KL}{Kullback-Leibler}
\abbreviation{RMSE}{Root-Mean-Square Error}
\abbreviation{Loess}{Local Polynomial Regression Fitting}
\abbreviation{NMH}{Normal Proposal Metropolis-Hastings}
\abbreviation{ESS}{Effective Sample Size}
\abbreviation{HPD}{Highest Posterior Density}
\abbreviation{CTMC}{Continuous Time Markov Chains}
\abbreviation{GRFS}{Global Rejection Free Sampler}
\abbreviation{ACF}{Autocorrelation Function}
\abbreviation{LBPS}{Local Bouncy Particle Sampler}
\abbreviation{MLE}{Maximum Likelihood Estimators}
\abbreviation{EM}{Expectation Maximization}
\abbreviation{PP}{Poisson Process}
\abbreviation{ARD}{Absolute Relative Difference}
\abbreviation{EIT}{Exact Invariance Test}
\abbreviation{PDMP}{Piecewise-deterministic Markov Process}
\abbreviation{NNPAAO}{Nearest Neighbour Pairwise Amino Acid Ordering}
\terminology{reversible}{reversible}
\terminology{leapfrog}{leapfrog}
\terminology{feature}{feature}
\terminology{featuretemplate}{feature template}
\terminology{finefeature}{fine feature}
\terminology{coarsefeature}{coarse feature}
\terminology{bivariatefeature}{bivariate feature}
\terminology{univariatefeature}{univariate feature}
\terminology{CLT}{Central Limit Theorem}

\let\oldepsilon\epsilon
\let\oldkappa\kappa
\let\oldDelta\Delta
\let\oldpi\pi
\let\oldtheta\theta

\notation[Space indexing the stochastic process]{tspace}{{\mathcal T}}
\notation*[A point in the index set of the stochastic process, i.e. a time point, or a point on a phylogenetic tree]{t}{t}

\notation[The state space of the process (at one time point)]{statespace}{{\mathcal X}}
\notation[A state of the process (at one time point)]{x}{x}
\notation{X}{X}
\notation[A process sample path]{xV}{\boldsymbol{x}}
\notation{XV}{\boldsymbol{X}}
\notation[Entry in the rate matrix]{q}{q}
\notation[Rate matrix]{Q}{Q}
\notation[Number of states in the CTMC]{nstates}{|{\mathcal X}|} 
\notation[Unordered pair of distinct states]{statespaceup}{{\mathcal X}^{\textrm{unordered,dist.}}}
\notation[Ordered pair of distinct states]{statespacep}{{\mathcal X}^{\textrm{distinct}}}
\notation{PQ}{\P_Q}
\notation{statio}{\oldpi}
\notation{ctmcsupport}{S}
\notation{totalLen}{\oldDelta}
\notation{eV}{\boldsymbol{e}}

\notation[Marginal transition probability of a CTMC]{ptrans}{p_{\textrm{trans}}}
\notation{fxlQDelta}{f_{\textrm{x}|\textrm{Q,}\oldDelta}} 
\notation{fxlQ}{f_{\textrm{x}|\textrm{Q}}} 
\notation[Initial distribution of the CTMC]{pini}{p_{\textrm{ini}}}
\notation[Emission density model]{femi}{f_{\textrm{emi}}}
\notation{fdylQ}{f_{\textrm{d,y}|\textrm{Q}}}
\notation{pdlQ}{p_{\textrm{d}|\textrm{Q}}}
\notation{pdlyQ}{p_{\textrm{d}|\textrm{y,Q}}}
\notation{fylQ}{f_{\textrm{y}|\textrm{Q}}}
\notation{fyld}{f_{\textrm{y}|\textrm{d}}}
\notation{fw}{f_{\textrm{w}}}
\notation{fwly}{f_{\textrm{w}|\textrm{y}}}
\notation{fwlzy}{f_{\textrm{w}|\textrm{z,y}}}
\notation{fwxyz}{f_{\textrm{w,x,y,z}}}
\notation{flga}{f_{\textrm{lga}}}
\notation{fnor}{f_{\textrm{nor}}}
\notation{flam}{f_{\lambda}}
\notation{fprop}{f_{\textrm{prop}}}
\notation{fzlwy}{f_{\textrm{z}|\textrm{w,y}}}

\notation[A list of observed time points]{tV}{\boldsymbol{t}}
\notation*[Inteval of time between two consecutive observed times]{Delta}{\oldDelta}
\notation[Inteval of time between two consecutive observed times]{branchlength}{\oldDelta}
\notation{DeltaV}{\boldsymbol{\oldDelta}}
\notation{obs}{\textrm{(obs)}}
\notation{y}{y}
\notation{Y}{Y}
\notation{yV}{\boldsymbol{y}}
\notation{YV}{\boldsymbol{Y}}
\notation{bi}{\boldsymbol{\textrm{bi}}}
\notation{uni}{\boldsymbol{\textrm{uni}}}
\notation*[The value of the latent process at one of the observation times]{d}{d}
\notation{D}{D}
\notation[A vector of the values of the latent process at all the observed times]{dV}{\boldsymbol{d}}
\notation{DV}{\boldsymbol{D}}  
\notation{ZV}{\boldsymbol{Z}}
\notation[Sufficient statistic of the latent process]{zV}{\boldsymbol{z}}
\notation[Sojourn time sufficient statistic vector (sum of holding times for each state)]{hV}{\boldsymbol{h}}
\notation{HV}{\boldsymbol{H}}
\notation[Sojourn time sufficient statistic for one state]{h}{h}
\notation[Transition counts (organized as a vector indexed by pairs of distinct states)]{cV}{\boldsymbol{c}}
\notation{CV}{\boldsymbol{C}}
\notation*[Transition count for one pair of distinct states)]{c}{c}
\notation[Initial state sufficient statistic vector]{nV}{\boldsymbol{n}}
\notation{NV}{\boldsymbol{N}}
\notation[Initial state sufficient statistic]{n}{n}
\notation{N}{N}

\notation[MCMC transition kerner]{mcmckernel}{T}

\notation{wV}{{\boldsymbol{w}}}
\notation{WV}{{\boldsymbol{W}}}
\notation{nr}{\textrm{(nr)}}
\notation{qnr}{q}
\notation{Qnr}{Q}
\notation{paramindex}{m}
\notation[Sufficient statistics of the CTMC]{suffV}{{\boldsymbol \varphi}}
\notation{suff}{\varphi}
\notation[Number of parameters (equivalently, dimensionality of the weight vector)]{nparams}{P} 
\notation{baseq}{\underset{\bar{}}{q}}
\notation*[Precision parameter of the prior on the weights]{kappa}{\oldkappa}
\notation{A}{A}

\notation[Number of leapfrog steps in HMC]{nleapfrogs}{L}
\notation*{epsilon}{\oldepsilon}
\notation{mcmciter}{i}
\notation{leapfrogiter}{j}
\notation[Negative log density targeted by the Hamiltonian Monte Carlo algorithm]{U}{U} 
\notation[Kinetic energy function used by the Hamiltonian Monte Carlo algorithm]{Kin}{K}
\notation[Moment auxiliary variable used by the Hamiltonian Monte Carlo algorithm]{mV}{{\boldsymbol{m}}}
\notation{HMCdist}{\textrm{HMC}}

\notation{hyperparam}{\gamma}
\notation{hyperparamspace}{\Gamma}
\notation{nadapt}{N_{\textrm{adapt}}}
\notation{BObeta}{\beta}

\notation[Set of all pairs of consecutive observations]{edgeset}{E} 
\notation[Set of all pairs of edges of the phylogenetic tree]{treeedgeset}{E} 
\notation[Pair of consecutive observations/edge of the phylogenetic tree]{e}{e}
\notation{nbranches}{|E|}
\notation[Index set for a discrete collection of times]{timeindexset}{V} 
\notation[Set of leaves and speciation points in the phylogenetic tree]{vertexset}{V}
\notation[Number of observed times in a time series]{nobs}{|V|}
\notation*[Index on an observation time, or, in phylogenetic, a node in the tree]{v}{v}
\notation[Number of sites (loci) in the phylogenetic setup]{K}{K}
\notation*[Site index in the phylogenetic setup]{k}{k}
\notation*[Root of the tree]{r}{r}
\notation{leaves}{V_{\textrm{leaves}}}
\notation{tree}{\boldsymbol{\tau}}

\notation{Xtree}{X}
\notation{XVtree}{\boldsymbol{X}}
\notation[A point on a phylogenetic tree]{ttree}{t}
\notation{Ytree}{Y}
\notation{YVtree}{\boldsymbol{Y}} 
\notation{treespace}{\mathcal{T}}
\notation{vtree}{v}
\notation{etree}{e}

\notation{sufftrans}{\phi}
\notation{suffstatio}{\psi}
\notation{sufftransV}{\boldsymbol{\phi}}
\notation{suffstatioV}{\boldsymbol{\psi}}
\notation{basetheta}{\underset{\bar{}}{\theta}}
\notation{basepi}{\underset{\bar{}}{\oldpi}}
\notation*{pi}{\oldpi}
\notation*{theta}{\oldtheta}
\notation{rev}{\textrm{(rev)}}
\notation{qrev}{q}
\notation{Qrev}{Q}
\notation{Nuni}{M_{\textrm{statio}}}
\notation{Nbi}{M_{\textrm{bi}}}
\notation{ZVrev}{\boldsymbol{Z}}

\notation{norm}{\textrm{(norm)}}
\notation{qnorm}{q}
\notation{Qnorm}{Q}
\notation{betaNorm}{\beta}

\notation{PL}{\P_{\boldsymbol{\lambda}}}
\notation{lamV}{{\boldsymbol{\lambda}}}
\notation{priorgamma}{f_{\textrm{gam}}}

\notation[Transition matrix derived from the rate matrix by adding self transitions, used by the uniformization method]{B}{B}
\notation[The maximum rate of departure among all states in the CTMC]{Qmax}{\bar q}
\notation[Number of jumps in the uniformization procedure]{J}{J}
\notation{basisvector}{\boldsymbol{e}}
\notation{cache}{\mathsf{cache}}
\notation{getAndCache}{\mathsf{get\_and\_cache}}
\notation{unifTransPrs}{\boldsymbol{p}}
\notation{unifTransPr}{p}

\notation[Number of non-zero entries across the features of all pairs of states]{s}{s}
\notation[Running time of the sum-product algorithm]{sumprodruntime}{T_0}
\notation{betaKenney}{\beta}
\notation[Expected maximum number of transition in a branch of the phylogenetic tree]{me}{M_{\textrm{e}}}

\notation{POLARITYSIZEFeat}{\textsc{PolaritySize}}
\notation{POLARITYFeat}{\textsc{Polarity}}
\notation{GTRFeat}{\textsc{Gtr}}
\notation{POLARITYSIZEGTRFeat}{\textsc{PolaritySizeGtr}}
\notation{SIZEFeat}{\textsc{Size}}

\notation{sizemap}{\textrm{size}}
\notation{polaritymap}{\textrm{polarity}}

\ShortHeadings{Analysis of high-dimensional CTMCs using the Local Bouncy Particle Sampler}{Tingting Zhao and Alexandre Bouchard-C\^{o}t\'{e}}
\firstpageno{1}

\begin{document}

\title{
	Analysis of high-dimensional Continuous Time Markov Chains using the Local Bouncy Particle Sampler}

\author{\name Tingting Zhao \email zhaott0416@gmail.com \\
       \addr College of Information and Computer Sciences\\
       University of Massachusetts Amherst\\
       Amherst, MA,  01003, USA
       \AND
       \name Alexandre \ Bouchard-C\^{o}t\'{e} \email bouchard@stat.ubc.ca \\
       \addr Department of Statistics\\
       University of British Columbia\\
       Vancouver, BC, V6T 1N4, Canada}

\editor{Zhihua Zhang}

\maketitle

\begin{abstract}
Sampling the parameters of high-dimensional \CTMCs\ is a challenging problem with important applications in many fields of applied statistics. In this work a recently proposed type of non-reversible rejection-free \MCMC\ sampler, the \BPS, is brought to bear to this problem. \BPS\ has demonstrated its favourable computational efficiency compared with state-of-the-art \MCMC\ algorithms, however to date applications to real-data scenario were scarce. An important aspect of practical implementation of \BPS\ is the simulation of event times. Default implementations use conservative thinning bounds. Such bounds can slow down the algorithm and limit the computational performance. Our paper develops an algorithm with exact analytical solution to the random event times in the context of \CTMCs. Our local version of \BPS\ algorithm takes advantage of the sparse structure in the target factor graph and we also provide a graph-theoretic tool for assessing the computational complexity of local BPS algorithms. 
\end{abstract}


\section{Introduction}
\CTMCs\ have widespread applications ranging from  chronic multi-state disease progression  \citep{chen1996markov, combescure2003assessment, saeedi2011priors, liu2015efficient} to phylogenetics \citep{yin2012continuous}.   However, the estimation of parameters in \CTMCs\ is a challenging problem when incomplete data observations are only available at a finite number of time points. This is the case in a wide range of applications, for analyzing censored survival data~\citep{kay}, for describing panel data under Markov assumptions~\citep{kalbfleisch1985analysis}, for characterizing multi-state disease progression~\citep{jackson2003multistate}, and for inferring evolutionary processes~\citep{Jukes1969JCModel, zhao2016bayesian} using biological sequences.

A (homogeneous) \CTMC\ is a continuous-time stochastic process taking values on a finite or countable set. The parameters involved in a \CTMC\ are used to characterize the transitions between states and the distributions of the intervals between two consecutive transitions. The parameters are organized into a rate matrix. If the sample paths have been completely observed continuously over a finite time interval, statistical inference is straightforward. However, a more typical situation is that only partial observations of the states on a finite number of time points are available. For high dimensional rate matrices, efficient posterior inference is challenging. In particular, despite several algorithmic advances \citep{moler2003nineteen}, matrix exponentiation, which is required to compute the marginal distributions of \CTMCs, is still computationally expensive. 

In the following, we will make use of a flexible framework to parameterize rate matrices \citep{zhao2016bayesian}, which subsumes much of the earlier parameterizations \citep{kimura1980,hky1985}. This previous work used off-the-shelf \AHMC\ methods and did not exploit the sparsity often found in the parameterization of high-dimensional rate matrices. 

In order to exploit sparsity, we make use of recently developed Monte Carlo schemes based on non-reversible \PDMPs. In particular, we build our samplers based on the non-reversible rejection-free dynamics proposed in \citet{peters2012rejection} in the physics literature and later developed for statistical applications in \citet{bouchard2018bouncy}. It has been shown by \citet{neal2004improving,sun2010improving,chen2013accelerating, bierkens2016non} that non-reversible \MCMC\ algorithms can outperform reversible \MCMC\ in terms of mixing rate and asymptotic performance. Related sampling schemes have been developed including continuous-time Monte Carlo algorithms and continuous-time Sequential Monte Carlo algorithms \citep{pakman2017stochastic,fearnhead2018piecewise}. \citet{bierkens2019zig} have developed a super-efficient sampling algorithm based on the zig-zag process under a big data context. But we focus on proof-of-concept applications of the \BPS\ algorithm in this work.  

In the \BPS\ algorithm,  the posterior samples of a variable of interest are continuously indexed by the position of a particle that moves along piecewise linear trajectories. When encountering a high energy barrier (low posterior density value), the particle is never rejected but instead the direction of its path is changed after a Newtonian collision. A key algorithmic requirement is to efficiently determine the bouncing time of the particle. Most existing work \citep{ vanetti2017piecewise, pakman2017stochastic, fearnhead2018piecewise} use conservative bounds from thinning algorithm of an inhomogeneous \PP\ to sample the collision time. Conservative bounds can lead to computational inefficiency of the algorithm. \citet{bouchard2018bouncy} has obtained analytical solutions to the collision times under certain simple scenarios such as Gaussian distribution. Thus, the \BPS\ can be highly efficient but does require application-specific work since the derivation of the collision time is case-specific. Therefore, we focus on demonstrating a concrete case study of how this can be achieved in the context of CTMCs and what potential benefits can be obtained.

The key contributions of this paper are as follows:

\begin{itemize}
\item Efficient algorithms to simulate the bouncing time for each factor of the factorized posterior density of \CTMCs\ to boost the computational efficiency. 
\item A novel sampler combining \HMC\ and \LBPS\ to efficiently sample from \CTMCs. This sampler is also of interest for sampling in other sparse factor graphs. 
\item A graph-theoretic tool to help select optimal sets of variables to be updated with \HMC\ and which ones to be updated with \LBPS, both for the special case of \CTMCs\ but also more generally in sparse factor graphs. 
\item A proof-of-concept application on protein evolution to demonstrate on real data the computational efficiency of \LBPS\ compared to state-of-the-art \HMC\ algorithms. 
\end{itemize}

The remaining of the paper is organized as follows. Section~\ref{sec:background} provides background on \CTMCs\ and on the \BPS\ algorithm. Section~\ref{sec:model} describes a novel model for \CTMCs\ used to demonstrate our computational methodology. Sections~\ref{sec:factorgraphcategory} and \ref{sec:algorithms} describe our key theoretical and methodological contributions. 
In Section~\ref{sec:experiments}, we compare our novel sampler to state-of-the-art methods. Finally, in Section~\ref{sec:real}, we provide results on protein data, and we discuss potential extensions of our work in Section~\ref{sec:extensions}.

\section{Problem setup and notations}\label{sec:background}
\subsection{\CTMCs\ notation}\label{sec:ctmc}
We first introduce some notation for \CTMCs. More background on \CTMCs\ can be found in \citet{norris1998markov,guttorp2018stochastic}. We use the same notation as \citet{zhao2016bayesian}. A (homogeneous) \CTMC\ is a continuous-time stochastic process $\{X(t): t\geqslant 0\}$ taking values on a finite or countable set $\statespace$. Throughout the paper, we assume $\statespace$ is finite and $\statespace = \{1, 2, \dots, \nstates\}$. 
Denote $\{X_n, n\geqslant 0\}$ as the sequence of states visited in the continuous time path $\{X(t)\}$, and let $A_n$ be the corresponding times when the state changes. A rate matrix $\Q$ indexed by $\statespace$ is used to describe the instantaneous transition rate for each pair of distinct states in $\statespace$. For example, $\q_{x, x'}$ represents the instantaneous rate between $x$ and $x'$, where $x\in\statespace, x'\in \statespace, x\ne x'$. The diagonal elements of $\Q$ are negative and enforce the constraint that each row sums to zero. The absolute values of the diagonal elements represent the rate parameter of an exponential distribution, used to characterize the waiting time spent on each state. We denote the initial state distribution as  $\pini(\cdot)$. Given the rate matrix $Q$, the transition probability matrix is $P_{\Q}(\Delta)=\exp({\Delta \Q})=\sum_{j=0}^{\infty} ({\Delta \Q}^j)/j!$, where $\Delta \ge 0$ corresponds to a time-step size. We use $\pi=(\pi_1, \pi_2, \ldots, \pi_{|\statespace|})$ to represent the stationary distribution of the \CTMC. 

\begin{figure}[t]
  \begin{center}
   \includegraphics[scale=0.25]
    {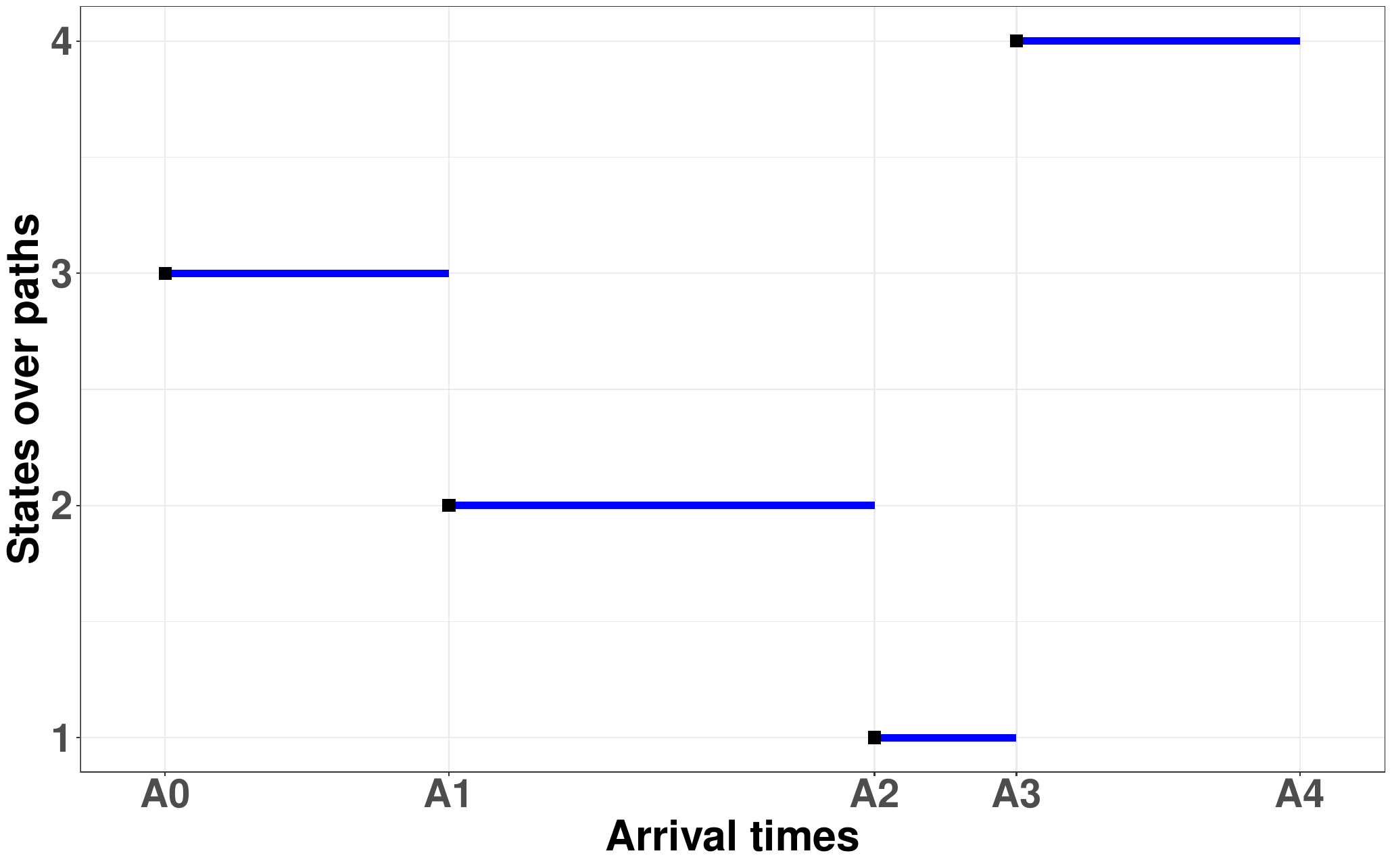}
  \end{center}
\caption{A fully observed realization of a single path of \CTMCs\ at arrival time points $A_1, A_2, A_3, A_4$ over a
state space $\{1, 2, 3, 4\}$.}
\label{fig:path}
\end{figure}

We use $\xV$ to represent $N$ fully observed \CTMC\ paths (example shown in Figure~\ref{fig:path}). We define $(\nV, \hV, \cV)\defeq (\nV(\xV), \hV(\xV), \cV(\xV))$ as the sufficient statistics of $N$ fully observed CTMC paths. For all $x\in\statespace$, denote $n_x \defeq n_x(\xV)\in \{0, 1, 2, \ldots \}$ as the number of paths started at state $x$ and $\nV(\xV) \defeq (n_1(\xV), n_2(\xV), \ldots, n_{\vert\statespace\vert}(\xV))$. Similarly, for all $x\in\statespace$, let $h_x\in [0, \infty)$ denote the total time spent in state $x$ and $\hV(\xV)\defeq (h_1(\xV), h_2(\xV), \ldots, h_{\vert\statespace\vert}(\xV)).$ For all distinct $(x, x')\in \statespace^{\textrm{distinct}}\defeq \{(x, x')\in \statespace^2: x\ne x'\}$, let $\c_{\x,\x'}$ denote the number of jumps from state $x$ to $x'$ and let $\cV(\xV)$ denote the vector of all transition counts organized according to some arbitrary fixed order of $\statespace^{\textrm{distinct}}$. For simplicity, we denote $\zV\defeq (\nV, \hV, \cV)$ and $\zV(\xV)\defeq (\nV(\xV), \hV(\xV), \cV(\xV))$. The density of $N$ fully observed paths $\xV$ from a \CTMC\ with rate matrix $Q$ over time interval length $\Delta$ is given by

\begin{eqnarray}\label{eq:fully-obs-density}
f_{x|Q, \Delta}(\xV| \Q, \totalLen) &\defeq& \left( \prod_{\x\in\statespace}  \pini(\x)^{\n_\x} \right) \left( \prod_{(\x,\x')\in\statespacep}  \q_{\x,\x'}^{\c_{\x,\x'}}\right) \\
&&\;\;\;\;\; \times \left( \prod_{\x\in\statespace} \exp\left(\h_\x \q_{\x,\x}\right) \right)\bold{1}[\xV\in S(\Delta)], \nonumber
\end{eqnarray} 
where $S(\Delta)$ denotes the set of \CTMC\ paths of total length $\Delta$.

Usually, these paths are only partially observed at a finite number of points $\tau_0, \tau_1, \ldots, \tau_n$ with states $y_0, y_1, \ldots$ and $y_n$. We use $\mathcal{Y}$ to represent the partially observed \CTMC\ path.  Assuming there is no  error in the observation of the states, the density over this single path is 
\begin{eqnarray}\label{eqn:single_path}
g_{\textrm{y}|\textrm{Q}}(\mathcal{Y}|Q)=\pini(y_{\tau_0})\prod_{k=1}^K \left(\exp(Q\Delta_k)\right)_{y_{\tau_{k-1}}, y_{\tau_k}},
\label{eq:partial_obs}
\end{eqnarray}
$\Delta_k=\tau_k-\tau_{k-1}$ is the length of the time interval between two consecutive observations. 

The question of interest is often to estimate the rate matrix given  partially observed paths. Pointwise evaluation of Equation~\ref{eq:partial_obs} can be done in $\mathcal{O}(|\statespace|^3)$ using diagonalization method to evaluate the matrix exponential.  We reduce the computational cost by introducing the substitution mapping auxiliary variables (instantiated via uniformization algorithms) as described in \citet{zhao2016bayesian}. With the augmented sufficient statistics,  the density of the paths is given by Equation~\ref{eq:fully-obs-density}.

\subsection{Bayesian \GLM\ parameterization}\label{sec:Bayesian GLM}

We describe here a sparse rate matrix parameterization built in the framework of  \citet{zhao2016bayesian}. This is useful since we propose Bayesian \GLM\ chain \GTR\ model in Section~\ref{sec:model} on the basis of Bayesian \GLM\ reversible rate matrix parameterization \citep{zhao2016bayesian}. We first describe the \GTR\ model and then review briefly the  Bayesian \GLM\ unnormalized reversible rate matrix parameterization. Finally, we review the Bayesian GLM rate matrix representation for a \GTR\ model since the notation will be needed to introduce our new sparse model in Section~\ref{sec:model}.

We introduce the set of unordered distinct pairs of states as  $\statespaceup \defeq \{\{\x, \x'\}\in \statespace^2 : \x \neq \x'\}$. Recall that in a \GTR\ model, $\Q$ is parameterized by the stationary distribution $\pi=(\pi_1, \pi_2, \ldots, \pi_{|\statespace|})$ and exchangeable parameters $\theta_{\{x,x'\}}$, where $\q_{\x,\x'} = \theta_{\{\x,\x'\}} \pi_{\x'}, x\ne x'$. In total, there are $p_2=|\statespace|(|\statespace|-1)/2$ exchangeable parameters $\theta_{x, x'}$ under reversibility. In the Bayesian \GLM\ unnormalized reversible rate matrix parameterization, we have:

\begin{eqnarray} \label{eq:trans-exp-fam}
\theta_{\{\x, \x'\}}(\wV^b) &\defeq& \exp\big\{\dotprod{\wV^b}{\sufftransV(\{\x,\x'\})}\big\},\label{eq:theta}   \\
\pi_x(\wV^u) &\defeq& \exp\big\{ \dotprod{\wV^u}{\suffstatioV(\x)} - \A(\wV^u)\big\} , \label{eq:statio-exp-fam} \\
\A(\wV^u) &\defeq& \log \sum_{\x \in \statespace} \exp\big\{ \dotprod{\wV^u}{\suffstatioV(\x)} \big\}, \label{eq:normalizing_constant}\\
\qrev^\rev_{\x, \x'}(\wV) &\defeq& \theta_{\{\x,\x'\}}\left(\wV^b\right) \pi_{\x'}\left(\wV^u\right), 
\label{eq:qxx}
\end{eqnarray}
The parameters of interest are $\wV$, where $\wV = 
\begin{pmatrix}
\wV^u\\
\wV^b
\end{pmatrix}.
$ In Equation~\ref{eq:theta}, we introduce bivariate feature functions $\sufftransV : \statespaceup \to \R^{p_2}$. In Equation~\ref{eq:statio-exp-fam}, we introduce univariate feature functions $\suffstatioV : \statespace \to \R^{p_1}$. We have $p_1+p_2=p$ and $\wV\in \R^{p}$. The weights related to the \emph{\univariatefeatures} $\suffstatioV$ are denoted as univariate weights $\wV^u$ and the weights related to the \emph{\bivariatefeatures} $\sufftransV$ are denoted as bivariate weights $\wV^b$.
 
The connection between Bayesian \GLM\ rate matrix representation and a \GTR\ model is discussed in details in Supplementary 8.1 and Supplementary 8.2 by \citet{zhao2016bayesian}. In Supplementary 8.1, \citet{zhao2016bayesian} have proved that the Bayesian \GLM\ can be used to represent any rate matrices under the \GTR\ parameterization equivalently. Here, we review this construction since our new sparse model in Section~\ref{sec:model} can be represented in a similar fashion. To represent the \GTR\ model in a Bayesian \GLM\ framework, if we have a \GTR\ rate matrix $Q$ with stationary distribution $\pi_x$ for any $x\in\statespace$ and $\sum_{x\in\statespace}\pi_x=1$, we can pick any reference state $x^*$ such that $$\pi_{x^*}(\wV)=\frac{1}{1+\sum_{x'\ne x^*:x'\in\statespace}\exp(\dotprod{\wV}{\phi(x')})},$$
and for other state $x'$,
$$\pi_{x'}(\wV)=\pi_{x^*}(\wV)\exp(\dotprod{\wV}{\phi(x')}).$$ 
Thus, for any stationary distribution $\pi_{x}$ of a GTR rate matrix $Q$, there exists a unique set of weights $\wV$ such that $\pi_x(\wV)=\pi_x$.

Similarly, we define a map $\eta$: $\statespaceup \to \{1, 2, \ldots, |\statespace|(|\statespace|-1)/2\}$ such that the $i$th element of \emph{\bivariatefeatures} $\sufftransV^{\textrm{gtr}} : \statespaceup \to \R^{p_2}$ is defined as: 

\begin{eqnarray}
\sufftransV_i^{\textrm{gtr}}(\{x, x'\}) \defeq \1(\eta(\{x, x'\})=i),
\label{eq:gtr_phi}
\end{eqnarray}
where the $i$th \feature\ $\sufftransV_i^{\textrm{gtr}}(\{\x, \x'\})$ is equal to one if and only if the pair of unordered states $\{x, x'\}$ is mapped to $i$th exchangeable parameter via $\eta(\{x, x'\})$. Under the definition of $\sufftransV^{\textrm{gtr}}$, for any $\theta_{x,x'}$ of a GTR rate matrix $Q$, there exists a unique set of weights such that $\theta_{x, x'}=\exp\left(\wV_{\eta(\{x, x'\})}\right)$.


\subsection{Bayesian inference}
Under the Bayesian \GLM\ rate matrix parameterization defined in Equation~\ref{eq:theta}-\ref{eq:qxx}, the parameters of interest are $\wV$, where $\wV = 
\begin{pmatrix}
\wV^u\\
\wV^b
\end{pmatrix}$. We place a prior on $\wV$ with density denoted by $g_{\textrm{w}}(\wV)$. Following \citet{zhao2016bayesian}, we use a Gaussian distribution with mean zero and precision $\kappa$ for $g_{\textrm{w}}(\wV)$.


We assume the observations are partially observed states $y_1, y_2, \ldots, y_k$ at a finite number of time points $\tau_0, \tau_1, \ldots, \tau_k$ on each sample path. To simplify the notations, we assume that $\tau_0, \tau_1, \ldots, \tau_k$ are fixed and known. The density over one single path is given in Equation~\ref{eq:partial_obs}. Under the Bayesian \GLM\ parameterization, we denote the density by $g_{\textrm{y}|\textrm{w}}(\mathcal{Y}|Q(\wV))$. Following Equation~\ref{eq:partial_obs}, we obtain:

\begin{eqnarray}\label{eqn:single_path_g}
g_{\textrm{y}|\textrm{w}}(\mathcal{Y}|Q(\wV))=\pini(y_{\tau_0})\prod_{k=1}^K \left(\exp(Q(\wV)\Delta_k)\right)_{y_{\tau_{k-1}}, y_{\tau_k}},
\label{eq:partial_obs_w}
\end{eqnarray}
Thus, the target posterior density is given by 

\begin{eqnarray}
 g_{\textrm{w}|\textrm{y}}(\wV|\mathcal{Y}) \propto g_{\textrm{w}}(\wV)g_{\textrm{y}|\textrm{w}}(\mathcal{Y}|Q(\wV)) =\exp(-\tilde U(\wV)),
\label{eq:posterior}
\end{eqnarray}
where $\tilde U(\wV)$ represents the negative logarithm of the unnormalized posterior density function. 

In Equation~\ref{eq:posterior}, the second term involves the computation of matrix exponentials. \citet{zhao2016bayesian} have argued that using \HMC\ directly on $\tilde U(\wV)$ leads to a high computational cost as a single gradient evaluation of Equation~\ref{eq:partial_obs_w} has a running time of $\Theta(\vert\chi\vert^5)$ \citep{zhao2016bayesian}. 
Applying other gradient-based MCMC methods such as BPS or LBPS would suffer from the same problem.

To circumvent this computational difficulty, we follow the auxiliary variable strategy of \citet{zhao2016bayesian}, where \emph{substitution mapping}  is used before each sampling step to augment the state space with (sufficient statistics of) full \CTMC\ sample paths. 
More precisely, given a pair of consecutive observed states separated by a time interval $\Delta$,  we simulate a full path conditionally on the end-points and time interval $\Delta$ according to  Equation~\ref{eq:fully-obs-density} using the cached uniformization techniques described in \citet{zhao2016bayesian}. 
After doing this for each pair of consecutive observed states, we obtain a vector of sufficient statistics $\zV \defeq (\nV(\xV), \hV(\xV), \cV(\xV))$ as defined in Section~\ref{sec:ctmc}. 

Following Equation~\ref{eq:fully-obs-density}, the negative unnormalized posterior log-density with augmented sufficient statistics $\zV$ is 
\begin{eqnarray}\label{eq:unnormalize_pointwise-eval}
 \U(\wV) \defeq  \U_{\zV}\left(\wV\right) &\defeq& 
      \frac{1}{2} \kappa \Vert \wV \Vert^2_2 
    - \sum_{\x \in \statespace} \h_\x  \qrev_{\x, \x}(\wV)
 \\
  &&\;\;- \sum_{(\x, \x') \in \statespacep} \c_{\x, \x'} \log\left(\qrev_{\x, \x'}\left(\wV\right)\right) - 
\sum_{\x \in \statespace}n_x\log(\pi_{x}(\wV))   \nonumber, \\
  &=&\frac{1}{2} \kappa \Vert \wV \Vert^2_2  
    + \sum_{(\x, \x') \in \statespacep}\h_\x \qrev_{\x, \x'}\left(\wV\right)\nonumber,
 \\
  &&\;\;- \sum_{(\x, \x') \in \statespacep} \c_{\x, \x'} \log\left(\qrev_{\x, \x'}\left(\wV\right)\right) - 
\sum_{\x \in \statespace}n_x\log(\pi_{x}(\wV))   \nonumber. 
. \end{eqnarray}

In  the remaining, we use the target density   $\exp(-U_{\zV}(\wV))$ instead of $\exp(-\tilde U(\wV))$. 
Appendix~\ref{sec:correctness_sampling_scheme} shows that $\exp(-U_{\zV}(\wV))$ admits the distribution of interest $\exp(-\tilde U(\wV))$ as a marginal, therefore there is no error involved in introducing the auxiliary variable $\zV$. 
Again we introduce this auxiliary variable to make sampling more computationally tractable as shown in Section~\ref{sec:algorithms}. 
 
\subsection{Background on \BPS\ and \LBPS}
\subsubsection{Basics of BPS}
The \BPS\ algorithm belongs to a type of emerging continuous-time non-reversible \MCMC\ sampling algorithms constructed from \PDMPs\ \citep{davis1984piecewise}. It was first proposed by \citet{peters2012rejection}, formalized, developed, and generalized by \citet{bouchard2018bouncy, vanetti2017piecewise}. 

\PDMPs\ alternate between deterministic trajectory segment each of a length determined by the first arrival of an inhomogeneous Poisson process, interspersed with random ``jumps.'' 
Three key components describe \PDMPs:
\begin{description}
\item[The deterministic dynamics:]  a system of differential equations that characterize the process' deterministic behaviour between jumps.
\item[The event rate:] a non-negative  function that determines the \emph{intensity} of jumps. As the process follows its deterministic trajectory, an inhomogenous Poisson process is defined by composing the state with the intensity function. The first arrival in this Poisson process determines the length of each deterministic segment. 
\item[The transition distribution:] given that a jump occurs, the transition kernel  is used to sample the next state given the current one. 
\end{description}
\BPS\ is a special case of \PDMPs\ with certain choices of the three aforementioned components.  We denote the state of the \BPS\ algorithm by $Y=(\boldsymbol{W}, \boldsymbol{V})$, which encodes the position and velocity of the particle. Let $\zeta(\wV, \boldsymbol{v})=\zeta(\boldsymbol{v})\zeta(\wV)$, where $\zeta(\boldsymbol{v})$ represents the standard multivariate Gaussian distribution and $\zeta(\wV)$ represents target the posterior distribution of interest. The \BPS\ is $\zeta(\wV, \boldsymbol{v})$-invariant. 

Equipped with this notation, we can now define the \BPS\ algorithm as a \PDMP\ with the following three components:

\begin{description}
\item[The deterministic dynamics:]
\begin{align*}
\frac{d\wV}{dt}=\boldsymbol{v}, \frac{d\boldsymbol{v}}{dt}=0.
\end{align*} 
\item[The event rate:] the intensity $\lambda(\wV, \boldsymbol{v})=\textrm{max}\{0, \dotprod{\nabla U(\wV)}{\boldsymbol{v}}\}$. 
\item[The transition distribution:] a Dirac centered at $(\wV, T_{\textrm{w}}(\boldsymbol{v}))$, where 

$$
T_{\textrm{w}}(\boldsymbol{v})=\boldsymbol{v}-2\frac{\dotprod{\nabla U(\boldsymbol{\wV})}{\boldsymbol{v}}}{\|\nabla U(\boldsymbol{\wV})\|^2}\nabla U(\boldsymbol{\wV}).
$$
\end{description}

To ensure the ergodicity of the Markov chain, ``refreshment events'' are also introduced.
We use independent refreshment, where the velocity is sampled according to $\zeta(\boldsymbol{v})$ (see \cite{vanetti2017piecewise} for alternatives). 
Refreshment events occur at times specified by independent and identically distributed exponential random variables with rate $\tau_{\textrm{ref}}$.

We summarize the \BPS\ sampler in Algorithm~\ref{alg:algo_bps}.

\begin{algorithm}[ht]
\caption{\BPS\ algorithm}
\label{alg:algo_bps}
\begin{algorithmic}[1]
{
\small
\STATE \textbf{Initialization:} \\
Initialize the particle position and velocity $\left( \wV^{(0)}, \boldsymbol{v}^{(0)}\right)$ arbitrarily on $\mathbb{R}^d\times\mathbb{R}^d$.\\
Set $T$ to a certain fixed trajectory length. 
\FOR{$i=1, 2, \ldots$,}
\STATE Sample the first arrival times $\tau_{\textrm{ref}}$ and $\tau_{\textrm{bounce}}$ of a \PP\ with intensity $\lambda^{\textrm{ref}}$ and $\lambda^{\textrm{bounce}}$ respectively, where $\lambda^{\textrm{bounce}}=\textrm{max}\left(\dotprod{\boldsymbol{v}^{(i-1)}}{\nabla U(\wV^{(i-1)}+\boldsymbol{v}^{(i-1)}t)}, 0\right)$ and the value of $\lambda^{\textrm{ref}}$ is pre-fixed. 
\STATE Set $\tau_i \leftarrow \textrm{min}(\tau_{\textrm{bounce}},  \tau_{\textrm{ref}})$.
\STATE Update the position of the particle via  $\wV^{(i)}\leftarrow \wV^{(i-1)}+\boldsymbol{v}^{(i-1)}\tau_i$.
\STATE If $\tau_i=\tau_{\textrm{ref}}$, sample the next velocity $\boldsymbol{v}^{(i)}\sim \mathcal{N}(0_d, I_d)$.
\STATE If $\tau_i=\tau_{\textrm{bounce}}$, obtain the next velocity $\boldsymbol{v}^{(i)}$ by applying the transition function $T_{\textrm{w}^{(i)}}\left(\boldsymbol{v^{(i-1)}}\right)$.
\STATE If $t_i=\sum_{j=1}^i\tau_j\geqslant T$, exit For Loop (line 2). 
\ENDFOR
}
\end{algorithmic}
\end{algorithm}
 
 \subsubsection{Basics of \LBPS}\label{sec:local_bps}
If the target posterior density $\zeta(\wV)$ can be represented as a product of positive factors in Equation~\ref{eq:factorizedlbps},
\begin{align}
\zeta(\wV)\propto \prod_{f\in\mathcal{F}}\gamma_f(N_f),
\label{eq:factorizedlbps}
\end{align} 
where $\mathcal{F}$ is the set of index for all factors in the target density and $N_f$ represents the subset of variables connected to factor $f$, then a ``local" version of \BPS\ referred to as \LBPS\ can be computationally efficient by taking advantage of the structural properties of the target density, especially if the target density has a \emph{strong sparsity} property described in definition~\ref{def:strong_sparsity}. 
When \LBPS\ is used, computationally cheaper refreshment scheme such as ``local refreshment"  \citep{bouchard2018bouncy} is often used by exploiting the structure of the factor graph. In the local refreshment scheme, one factor $\gamma_f$ is chosen uniformly at random first and only the components of $\boldsymbol{v}$ for variables in $N_f$ are resampled. The candidate collision time is recomputed only for the extended neighbour factors $\gamma_{f'}$ of $\gamma_f$, where $N_{f'}\cap N_f\ne \emptyset$.

Now we present a brief description of \LBPS\ algorithm and we will see how sparsity plays an important role in improving the computational efficiency. Detailed description about an efficient implementation of \LBPS\ via the priority queue can be found in \citet{bouchard2018bouncy}. If the target unnormalized posterior density can be factorized according to Equation~\ref{eq:factorizedlbps}, its associated energy function is 
\begin{eqnarray}
U(\wV) = \sum_{f\in\mathcal{F}} U_f(N_f),
\label{eq:energy_function}
\end{eqnarray}
where $U_f(N_f)=-\log(\gamma_f(N_f)).$ We define the local intensity $\lambda_f(\wV, \boldsymbol{v})$ and local transition function $T_{\textrm{w}}^f(\boldsymbol{v})$ for factor $\gamma_f$ as:

\begin{eqnarray}
\lambda_f(\wV, \boldsymbol{v})=\textrm{max}\{0, \dotprod{\nabla U_f(\wV)}{\boldsymbol{v}}\}\\
T_{\textrm{w}}^f(\boldsymbol{v})=\boldsymbol{v}-2\frac{\dotprod{\nabla U_f(\boldsymbol{\wV})}{\boldsymbol{v}}}{\|\nabla U_f(\boldsymbol{\wV})\|^2}\nabla U_f(\boldsymbol{\wV}).
\label{eq:velocity_update}
\end{eqnarray}
Note that for variables that are not the neighbour variables for factor $\gamma_f$, $T_{\textrm{w}}^f(\boldsymbol{v})_k = \boldsymbol{v}_k$. In \LBPS, the next collision time is the first arrival time of a \PP\ with intensity $\lambda(\wV, \boldsymbol{v})=\sum_{f\in\mathcal{F}}\lambda_f(\wV, \boldsymbol{v}).$ We can sample the first arrival time using the methods described in \cite{bouchard2018bouncy} (see Section~\ref{sec:solution_time} for specific examples for the models explored in this paper). We sample $\tau_f$ for each factor $\gamma_f$ from a \PP\ with intensity $\lambda_f(\wV, \boldsymbol{v})=\textrm{max}\{0, \dotprod{\nabla U_f(\wV)}{\boldsymbol{v}}\}$. The first arrival time for \PP\ with intensity $\lambda(\wV, \boldsymbol{v})$ is $\tau=\underset{f\in \mathcal{F}}{\textrm{min}} \,\, \tau_f$.  Once a bounce event takes place, \LBPS\ only needs to manipulate a subset of the variables and factors. To be specific, if we denote $f^*$ as the factor index such that $\tau_{f^*}=\underset{f\in \mathcal{F}} {\textrm{min}} \,\, \tau_f$, we use $\gamma_{f^*}$ to denote the collision factor. Following the priority queue implementation described in \citet{bouchard2018bouncy}, \LBPS\ will update the position of neighbour variables for the collision factor $\gamma_{f^*}$ and then apply the corresponding local transition function $T_{\textrm{w}}^{f^*}(\boldsymbol{v})$ to update the velocity. Next, the algorithm will sample the candidate bounce time of the next event for all the extended neighbour factors $\gamma_{f'}$ for factor $\gamma_{f^*}$ such that $N_{f'}\cap N_{f^*}\ne\emptyset$. If strong sparsity (defined in Section~\ref{sec:strong_spartisity}) is satisfied for a family of factor graphs, then the number of neighbour variables for the collision factor grows much slower than the dimension of the parameter space and is  negligible compared to the dimension of the parameter space. The number of operations needed to update the position of the neighbour variables is negligible. When computing the candidate collision time for the extended neighbour factors, the number of factors involved is also negligible compared to the total number of factors in the factor graph, which is desirable in \LBPS. We need to point out that although \LBPS\ manipulates a subset of variables and factors, each local bounce will lead to changes in all the variables instead of just the neighbour variables connected with the current collision factor. 

\section{Proposed Bayesian \GLM\ chain \GTR\ model} \label{sec:model}
In Section~\ref{sec:Bayesian GLM}, we introduced a function  
$\eta(\{x, x'\})$ which embeds state transitions into the integer. We will call this $\eta$ an ``ordering function''. Here we use such ordering function to model the exchangeable rates between each pair of states through the current pair and its nearest neighbour pair. We would like to find an order of each pair of distinctive states that is biologically meaningful so that parameters that are neighbours in this ordering can share statistical strength.  Take protein evolution as an example. Most frequent amino acid exchanges take place between residues with similar physicochemical properties. Under the Bayesian \GLM\ chain \GTR\ model we proposed below, we assume that mutations between two pairs of amino acids sharing one common state are expected to have similar exchangeable rates if the unshared distinctive states between the two pairs have similar physicochemical properties. We would like to find an ordering  $\eta(\{x, x'\})$ such that amino acid pairs that are nearest neighbours
share the most similar physicochemical properties given one common state. The \NNPAAO\ Algorithm (described in Appendix F) proposed in Section~\ref{sec:real} provides an ordering of amino acid pairs that satisfies such property. We denote it as $\eta_{\textrm{dist}}(\{x, x'\})$ since the order is determined according to the Euclidean distance between amino acids defined in Equation~\ref{eq:distance}. Given this ordering, neighbour pairs share more biological similarities than non-neighbour pairs. 

In this paper, based on the intuition of $\eta_{\textrm{dist}}(\{x, x'\})$, we define a novel Bayesian \GLM\ \textbf{chain} \GTR\ model with feature function

\begin{eqnarray}
\sufftransV^{\textrm{chain}}_i(\{x, x'\}) 
\defeq \1(\eta(\{x, x'\})\in \{i, i+1\}),
\label{eq:chain_phi}
\end{eqnarray}
where $i=1, 2, \ldots, |\statespace|(|\statespace|-1)/2.$ If we set $\eta((x, x'))=\eta_{\textrm{dist}}(\{x, x'\})$, the intuition of the Bayesian \GLM\ chain \GTR\ model is that the value of each exchangeable parameter $\theta_{\{x, x'\}}$ depend on $\{x, x'\}$ and its neighbour pair denoted as $\{x{''}, x{'''}\}$. Its neighbour pair satisfies that $\vert\eta(\{x{''}, x{'''}\})-\eta(\{x, x'\})\vert=1.$ Equivalently, we can define the  Bayesian chain \GTR\ model through the exchangeable parameters $\theta_{\{\x, \x'\}}(\wV^b)$ by plugging in the definition of $\sufftransV_i^{\textrm{chain}}(\{x, x'\})$ in Equation~\ref{eq:chain_phi}:

\begin{eqnarray}
\theta_{\{\x, \x'\}}(\wV^b) &=& \exp\big\{\dotprod{\wV^b}{\sufftransV^{\textrm{chain}}(\{\x,\x'\})}\big\}\nonumber\\
&=&\begin{cases}
\exp\left(\wV^b_{\eta(\{x, x'\})}+\wV^b_{\eta(\{x, x'\})-1}\right), &\text{if} \,\, \eta(\{x, x'\})=2, 3, \ldots, (|\statespace|(|\statespace|-1)/2),\nonumber\\
\exp\left(\wV^b_{\eta(\{x, x'\})}\right), &\text{if}\,\, \eta(\{x, x'\})=1.
\end{cases}
\end{eqnarray}
For simplicity, we focus on the simple example of chains, but this model could be generalized to other sparse graphs. We provide a general characterization of the running time analysis for both arbitrary factor graphs and factor graphs with sparse structure in Section~\ref{sec:running_time_lbps_general}. The chain \GTR\ model is general in the sense that it is able to represent any rate matrices. We prove this in  Appendix A.

\section{Characterization of factor graphs where \LBPS\ is efficient}\label{sec:factorgraphcategory}
\subsection{Strong sparsity of factor graphs}\label{sec:strong_spartisity}
A factor graph is a bipartite graph used to represent the factorization of a function, often a density function (as in \citet{bouchard2018bouncy}), or a conditional density function (as in Section~\ref{sec:alternation}). 
We will use factor graphs to define a non-standard notion of sparsity, \emph{strong sparsity}, that implies computational efficiency guarantees for the \LBPS\ algorithm.

\begin{figure}[ht] 
  \begin{center}
   \includegraphics[scale=0.23]
    {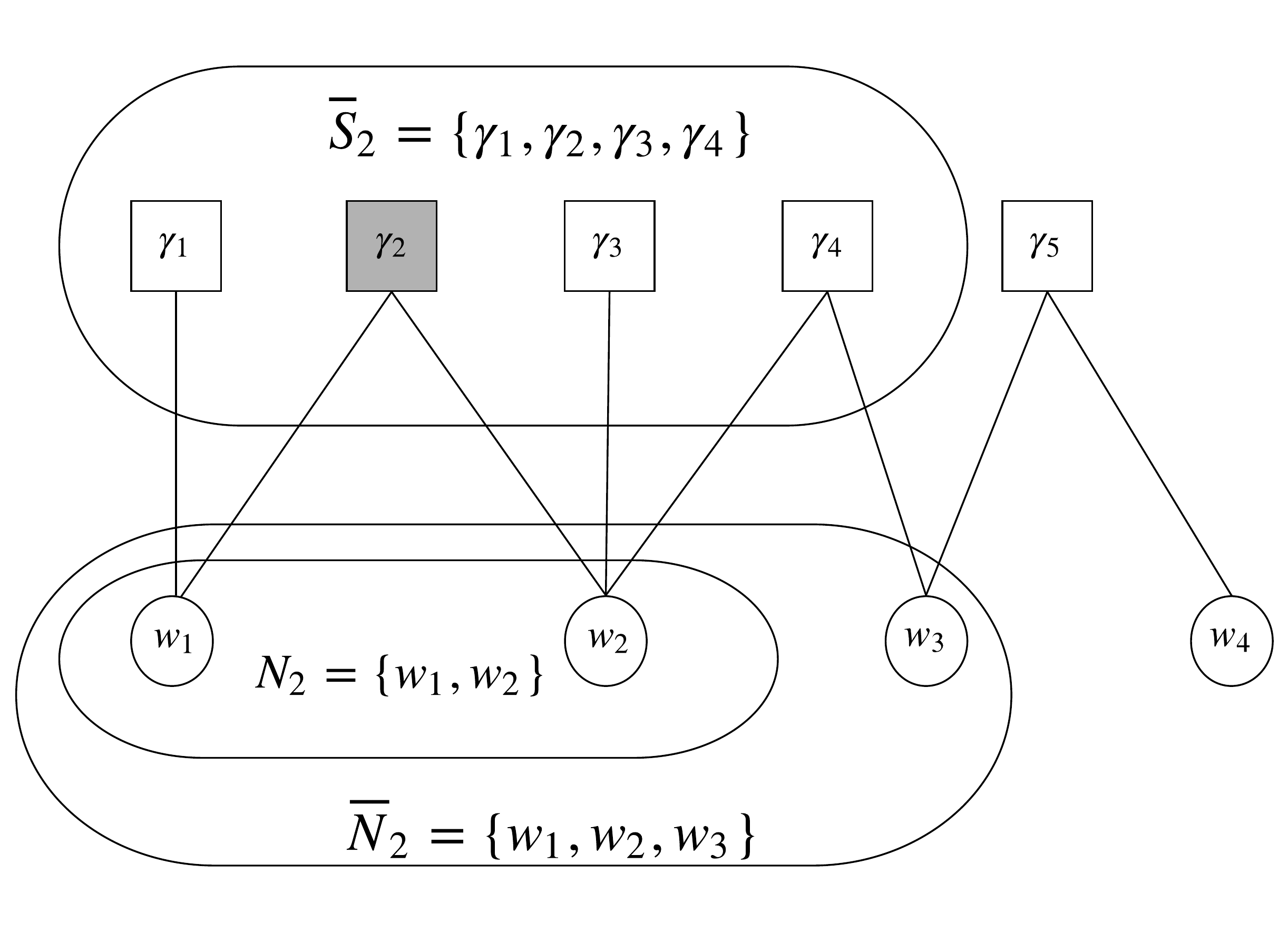}
  \end{center}
\caption{For factor $\gamma_2$, its neighbour variables $N_2=\{w_1, w_2\}$. Its extended neighbour variables $\overline{N}_2=\{w_1, w_2, w_3\}$. Its extended neighbour factors $\overline{S}_2=\{\gamma_1, \gamma_2, \gamma_3, \gamma_4\}$.}
\label{fig:factor_graph_example}
\end{figure}

We start by formalizing the notion of factor graph and some associated definitions: 


\begin{definition}
Given a factorization of a function
$$f(\wV)=\prod_{f=1}^{m}\gamma_f(N_{f}),$$
where $N_f \subset\{w_1, w_2, \ldots, w_p\}$ denotes \textbf{neighbour variables} for factor $\gamma_f$, 
the \textbf{factor graph} $G=(\wV, \Gamma, E)$ consists of a set of variables $\wV=\{w_1, w_2, \ldots, w_p\}$, a set of factors $\Gamma=\{\gamma_1, \gamma_2, \ldots, \gamma_m\}$ and a set of edges E defined as follows. We place an undirected edge between factor $\gamma_f$ and variable $w_k$ if and only if $w_k\in N_f$.  The \textbf{neighbour factors} for variable $w_k$ is $S_k\defeq\{\gamma_f:w_k\in N_f\}$. The \textbf{extended neighbour variables} for factor $\gamma_f$ is $\overline{N}_f\defeq\{w_k: w_k\in N_{f'}\,\, \textrm{such that}\,\, N_{f'}\cap N_f\ne \emptyset\}$. The \textbf{extended neighbour factors} for factor $\gamma_f$ is $\overline{S}_f\defeq\{\gamma_{f'}: N_{f'}\cap N_{f}\ne\emptyset \}.$
\label{def:neighbourhood}
\end{definition}
In Figure~\ref{fig:factor_graph_example}, we use the following factor graph to illustrate definition~\ref{def:neighbourhood}:
\begin{eqnarray*}
f(\wV)=\gamma_1(w_1)\gamma_2(w_1, w_2)\gamma_3(w_2)\gamma_4(w_2, w_3)\gamma_5(w_3, w_4).
\label{eq:factor_graph_eg}
\end{eqnarray*}

Next, we define the notion of \textbf{strong sparsity}, which applies to factor graph families in which the number of factors $m$ and variables $p$ goes to infinity. 
\begin{definition}
	A family of factor graphs is \textbf{strongly sparse} if  
$\underset{f}{\max}\,\,\vert  \overline{S}_f \vert = o(m),$ for $1\leqslant k \leqslant p,$ and $\underset{f}{\max}\,\, \vert  \overline{N}_f \vert = o(p)$, for $1\leqslant f\leqslant m$.
\label{def:strong_sparsity}
\end{definition}

Strong sparsity ensures that for any factor in the factor graph, the cardinality of its extended neighbour factors and extended neighbour variables grows asymptotically slower than the dimension of the parameter space and is ultimately negligible compared to the dimension of the parameter space and the total number of factors in the factor graph. We will show later that the factor graph under our proposed \CTMC\ \GLM\ model  sampling scheme satisfies the strong sparsity property. 

\subsection{Running time analysis of \LBPS\ for factor graphs}\label{sec:running_time_lbps_general}
Recent work \citep{deligiannidis2018randomized, andrieu2018hypercoercivity} has analyzed the scaling limits of samplers constructed from \PDMPs\ including \BPS\ and Zig-zag processes. However, we are not aware of existing analyses of the running time of the \LBPS\ algorithm. Here we only partially fill this gap. Specifically, we provide an analysis of the running time for a fixed trajectory length. This leaves open the scaling property of the \ESS\ for \LBPS\ ran for a fixed trajectory length. Previous work has shown that the \ESS\ of the \BPS\ algorithm is proportional to the trajectory length \citep{deligiannidis2018randomized}, but this result is for the global \BPS\ algorithm. 

Despite this limitation, our per-iteration cost analysis already reveals an interesting running time gap between \LBPS\ executed on strongly sparse graphs versus general factor graphs. We stress that the emphasis here is not to build a comprehensive scaling limit for the purpose of comparing \LBPS\ to other MCMC algorithms, but rather to guide the choice of factorization to use when constructing an \LBPS\ algorithm. Specifically, we show in Section~\ref{sec:alternation} that by using a blocking strategy, we can increase the sparsity of \LBPS's factor graph, by going from the factorization of the joint to the factorization of a conditional.

We first introduce some notations needed in the analysis. Denote $\left\vert\overline{N}_*\right\vert=\underset{f}{\mbox{max}}\,\,\left\vert \overline{N}_f\right\vert, \left\vert N_*\right\vert=\underset{f}{\mbox{max}}\,\,\left\vert N_f\right\vert$ and $\left\vert\overline{S}_*\right\vert=\underset{f}{\mbox{max}}\,\,\left\vert\overline{S}_f\right\vert$. Since $N_f\subset \overline{N}_f$, we have $\left\vert N_f\right\vert\leqslant \left\vert \overline{N}_f\right\vert$ and $\left\vert N_*\right\vert\leqslant \left\vert\overline{N}_*\right\vert$. Denote the cost for computing the collision time for a factor $\gamma_f$ as $c_f$ and $c_*=\underset{f}{\mbox{max}}\,\, c_f$. Let $c_{U_f}$ denote the running time for computing $\nabla U_f(\wV)$ and $c_{U_{*}}=\underset{f}{\mbox{max}}\,\,c_{U_f}$. Now, we analyze the running time of \LBPS\ via a priority queue implementation \citep{bouchard2018bouncy} for factor graph $G=(\wV, \Gamma, E)$. 

\begin{enumerate}
\item Compute the collision time for all factors and build a priority queue: $\mathcal{O}(mc_{*}+m\log m)$.
\item 
\begin{enumerate}
	\item If a collision takes place, 	\begin{enumerate}
		\item Update the position of extended neighbour variables: $\mathcal{O}(\left\vert \overline{N}_{*} \right\vert)$.
		\item Update the velocity of neighbour variables according to Equation~\ref{eq:velocity_update}: $\mathcal{O}(\vert \overline{N}_{*}\vert+c_{U_{*}})$.
		\item Add new samples to the trajectory list $L_k$: $\mathcal{O}(\vert N_{*}\vert)$. We use $L_k$ to denote a list of triplets $\left( w_k^{(i)}, v_k^{(i)}, t_k^{(i)}\right)$, with $ w_k^{(0)}$ and $v_k^{(0)}$ representing the initial position, and velocity and $t_k^{(0)}=0$. For $i>0$, $v_k^{(i)}$ represents the velocity after the $i$th collision or refreshment event and $t_k^{(i)}$ represents the time for the $i$th event. 
		\item Compute the collision time for extended neighbour factors: $\mathcal{O}\left( \left\vert \overline{S}_*   \right\vert c_{*} \right)$.
		\item Insert the extended neighbour factors and its collision time into the priority queue: $\mathcal{O}(\log m)$. The elements of the priority queue are stored according to an increasing order of the collision time.  
	\end{enumerate}
\item If a refreshment event takes place and a local refreshment scheme (described in Section~\ref{sec:local_bps}) is used:
	\begin{enumerate}
	\item Pick a factor uniformly at random and refresh the velocity: $\mathcal{O}(\vert N_{*}\vert)$.
	\item Update the position of the neighbour variables of the selected factor, compute the collision time for the extended neighbour factors and update the collision time in the priority queue: $\mathcal{O}(\vert \overline{N}_{*}\vert+ \left\vert \overline{S}_* \right\vert c_{*} +\left\vert \overline{S}_* \right\vert\log m)$. 
	\item Sample the next refreshment time $\mathcal{O}(1)$. 
	\end{enumerate}
\end{enumerate}
\end{enumerate}
Thus, assuming there are $J_1$ collision events and $J_2$ refreshment events when the particle travels along a fixed trajectory length, the total running time for \LBPS\ is: 
$$
\mathcal{O}\bigg(mc_*+ J_1
\left(  \left\vert\overline{N}_*\right\vert +c_{U_{*}}+
\left\vert\overline{S}_*\right\vert c_*+\log m\right)  +J_2(\vert \overline{N}_{*}\vert+ \left\vert \overline{S}_* \right\vert c_{*} +\left\vert \overline{S}_* \right\vert\log m) \bigg).
$$
For a factor graph with no sparsity, the total running time for \LBPS\ is 
$$
\mathcal{O}\bigg( J_1
\left(  p+c_{U_{*}}+
mc_*+\log m\right)  +J_2(p+mc_{*}+m\log m) \bigg).
$$
For a factor graph with strong sparsity, the total running time for \LBPS\ is 
\begin{eqnarray}
\mathcal{O}\bigg(mc_*+ J_1
\left(  p^{\alpha_1}+c_{U_{*}}+
m^{\alpha_2}c_*+\log m\right)  +J_2(p^{\alpha_1}+m^{\alpha_2}c_{*}+m^{\alpha_2}\log m) \bigg),
\label{eq:compute_cost_sparsity}
\end{eqnarray}
where $0\leqslant\alpha_1<1, 0\leqslant\alpha_2<1.$ 

If $m$ and $J_1$ are the dominant terms, as typical in applications, this yields a reduction in per-iteration cost from $\mathcal{O}(J_1 m)$ down to $\mathcal{O}(J_1 m^{\alpha_2})$ when moving from a general factor graph to a strongly sparse one.

\section{Methodology}\label{sec:algorithms} 

\subsection{Achieving strong sparsity through \LBPS-\HMC\ alternation}\label{sec:alternation}\label{sec:factor_graph}

In this section, we first show that applying \LBPS\ naively to the \GLM\ \CTMC\ models introduced in Section~\ref{sec:Bayesian GLM} would lead to a factor graph which does not satisfy strong sparsity, and hence to an inefficient \LBPS\ algorithm. 
To address this issue, we will introduce a sampling strategy that alternates between two moves: one which samples a subset of variables based on \LBPS, while conditioning on the rest, followed by one move which samples  a small set of more connected variables based on \HMC\ while conditioning on the rest. 

To motivate the \LBPS-\HMC\ alternation algorithm, let us return to the problem of posterior inference of the parameters of \CTMC\ \GLM\ models. 
In any such models, we can decompose the augmented negative log density $U(\wV)$ (Equation~\ref{eq:unnormalize_pointwise-eval}) into a sum of factors. We first introduce notations to group these factors  into the following categories: 
\begin{description}
	\item [Normal factor:] $\frac{1}{2} \kappa \Vert \wV \Vert^2_2,$ 
	\item[Sojourn time factor:] $H_{x, x'}\defeq\h_\x \qrev_{\x, \x'}\left(\wV\right),$ for all $(\x, \x') \in \statespacep$,
	\item[Transition count factor:] $C_{x, x'}\defeq -\c_{\x, \x'} \log\left(\qrev_{\x, \x'}\left(\wV\right)\right)$, for all $(\x, \x') \in \statespacep$,
	\item[Initial count factor:] $\boldsymbol{\pi}_x\defeq-n_x\log(\pi_{x}(\wV))$, for all $\x \in \statespace$,  with $\boldsymbol{\pi}_x$ representing the initial count factor but $\pi_x(\wV)$ denotes the stationary distribution for state $x$ given $\wV$. 
\end{description} 

\begin{figure}[ht] 
	\begin{center}
		\includegraphics[scale=0.45]
		{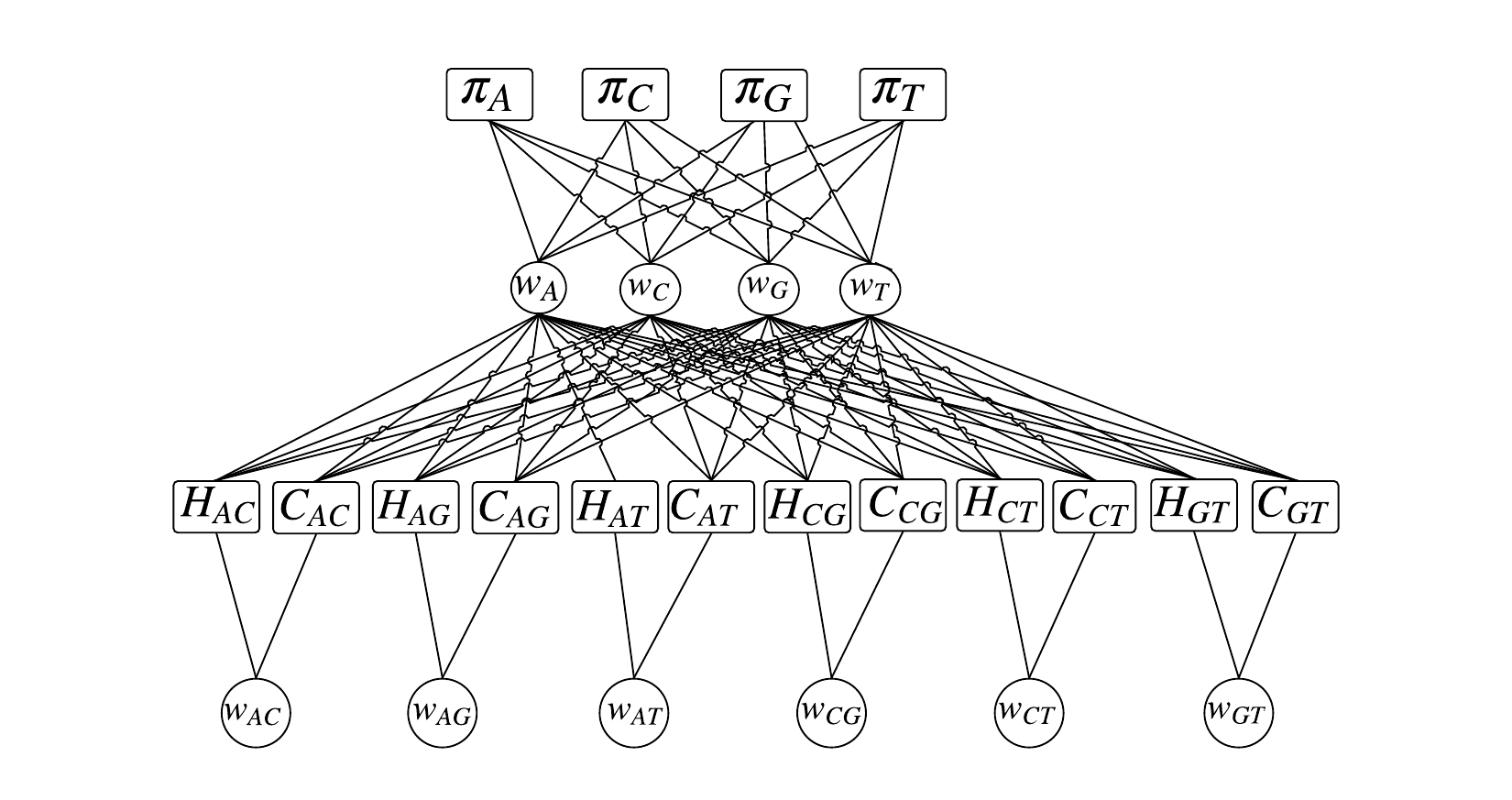}
	\end{center}
	\caption{Factor graph corresponding to the full posterior of the parameters of the  \GLM\ representation of \GTR\ model (shown here for DNA data). Top row shows the initial count factors, second row, stationary distribution parameters, third row, transition and sojourn time factors ($C$ and $H$ respectively), last row, exchangeable parameters. The graph is not strongly sparse.  To make the figure simpler, we have omitted redundant factors which do not affect the strong sparsity argument (those corresponding to the normal prior, the other half of the  transition count factors and  sojourn time factors for the symmetric pair of states $(x', x)$ such as $H_{CA}, C_{CA}, H_{GA}, C_{GA}, \ldots$. 
	}
	\label{fig:factor_graph_all_connected}
\end{figure}

Consider first the \GTR\ version of this model. The factor graph of the full posterior distribution is shown in Figure~\ref{fig:factor_graph_all_connected}. It is the sparsest possible \CTMC\ \GLM\ model, in the sense that there is no parameter sharing among the exchangeable nor stationary parameters, yet, it can be readily seen that this graph is not strongly sparse. 
Indeed, this graph has the property that for any factor $\gamma$, the extended neighbourhood of $\gamma$ includes all other factors since they all share common neighbour variables $\wV^u$. Hence the graph is not strongly sparse and \LBPS\ does not enjoy the computational savings described in Section~\ref{sec:running_time_lbps_general}.

To address this issue, instead of using \LBPS\ on the full posterior distribution, we only apply \LBPS\  to a subset of the parameters while conditioning on the others. Within one \LBPS\ round, only a subset of the variables are updated while keeping the other ones fixed. 
This strategy can be described as an ``\LBPS-within-Gibbs'' method, in parallel to the terminology ``Metropolis-within-Gibbs'' (see e.g.\ \citet{diggle_model-based_1998}). 
We still maintain invariance with respect to the correct distribution since we alternate between \LBPS\ and an \HMC\ move sampling the variables not updated by \LBPS. As we will show shortly, in the context of \CTMC\ \GLM\ models, it is advantageous to fix the parameters $\wV^{u}$ during the \LBPS\ phase of the alternation. The proposed sampling scheme is summarized in Algorithm~\ref{alg:algo_lbps}, and its invariance is established formally in Appendix~\ref{sec:correctness_sampling_scheme}.

\begin{algorithm}[ht]
	\caption{Proposed sampling scheme \LBPS-\HMC\ for \CTMCs}
	\label{alg:algo_lbps}
	\begin{algorithmic}[1]
		{
			\small
			\STATE \textbf{Initialization:} \\
			Initialize weights $\wV^{0}=(\wV^{u}, \wV^{b})$ from $N(0, 1)$, where $\wV^u$ represents the weights corresponding to the \emph{\univariatefeatures} to compute the stationary distribution and $\wV^b$ represents the weights associated with \emph{\bivariatefeatures} to compute the exchangeable parameters. 
			\FOR{$i=1, 2, \ldots, N$,}
			\STATE Compute the rate matrix $Q$ given $\wV^{(i-1)}$ under Bayesian \GLM\ rate matrices. 
			\STATE Use an end-point sampler as in \cite{zhao2016bayesian} to simulate a path given time interval $\Delta_{e}$ with two consecutive observations observed at time points $e=(t, t+1)$ of the time series according to Equation~\ref{eq:fully-obs-density} using the cached uniformization technique.
			\STATE Compute the aggregate sum of the sufficient statistics of all time series obtained in Step 4. 
			\STATE Update univariate weights $\wV^u$ for the stationary distribution via \HMC: 
			$$\left(\wV^{u}\right)^{(i+1)}|\left(\wV^{u}, \wV^{b}\right)^{(i)}, Z^{(i)}\sim \textrm{HMC}(\cdot \,|\left(\wV^{u}, \wV^{b}\right)^{(i)}, Z^{(i)}, L, \epsilon),$$
			where $L$ and $\epsilon$ are tuning parameters representing the number of leapfrog jumps and step size in \HMC. Recall that $Z\defeq (N, H, C)$ (defined in Section~\ref{sec:ctmc}) denotes the sufficient statistics of the augmented CTMC paths. Function evaluation and gradient calculation required by \HMC\ are described in Equation~\ref{eq:uni_function_evaluation} and Equation~\ref{eq:gradient_hmc_uni} in Appendix B. 
			\STATE Update bivariate weights $\wV^{b}$ used to compute the exchangeable parameters:
			$$\left(\wV^{b}\right)^{(i+1)}|\left(\wV^{u}\right)^{(i+1)}, \left(\wV^{b}\right)^{(i)},  Z^{(i)}\sim \textrm{LBPS}(\cdot \,|\left(\left(\wV^{u}\right)^{(i+1)}, \left(\wV^{b}\right)^{(i)}\right), Z^{(i)}, T),$$
			where $T$ is the tuning parameter in \LBPS\ representing the fixed length of the trajectory. 
			
			\ENDFOR
			
		}
	\end{algorithmic}
\end{algorithm}

When performing \LBPS\ on a subset of the latent variables, the variables that are temporarily fixed can be removed from the factor graph, as their values do not change in this segment of \LBPS\ trajectory. 
The power of our \LBPS-\HMC\ alternation method comes from the fact that in certain situations we only need to fix a small number of variables in order to gain strong sparsity.

\begin{figure}[t]
	\begin{center}
		\includegraphics[scale=0.3]
		{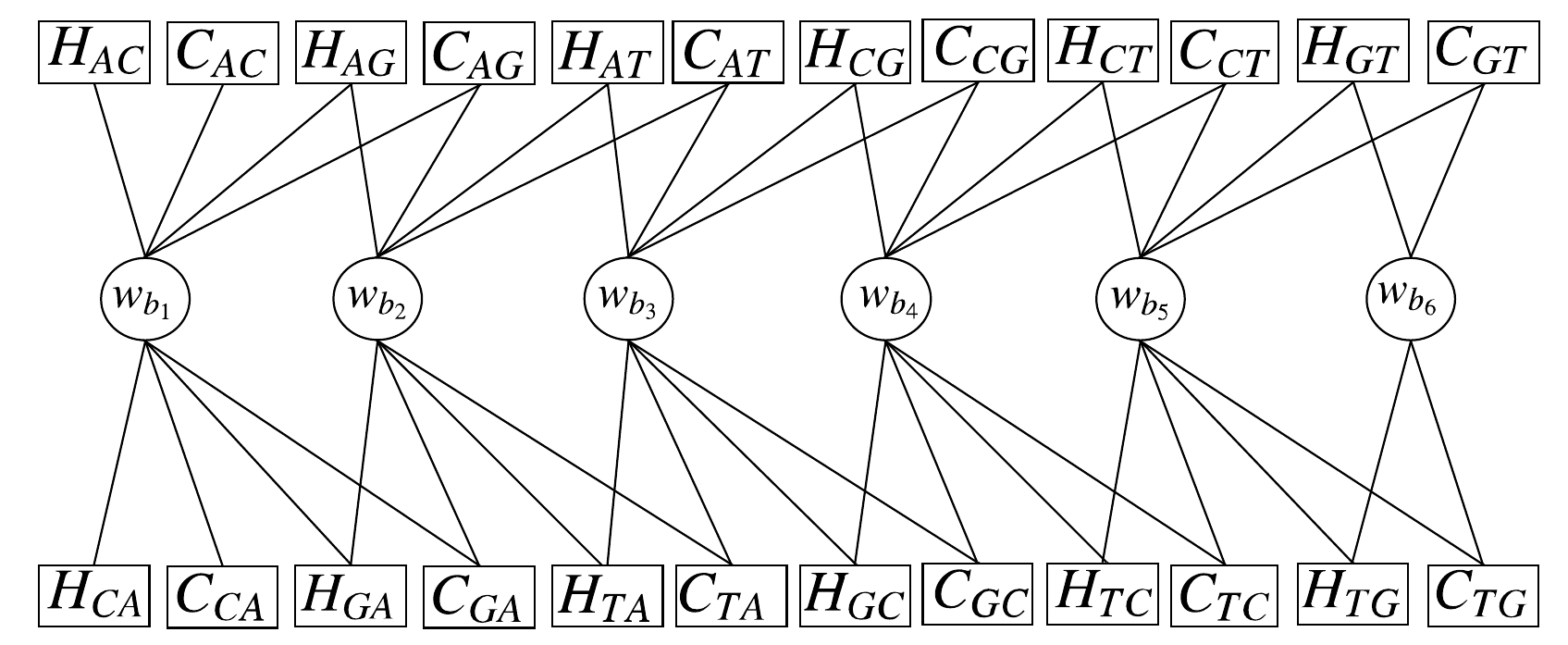}
	\end{center}
	\caption{Factor graphs involved in the \LBPS\ phase of our \LBPS-\HMC\ alternation algorithm applied to the chain \GTR\ model. This is the factor graph corresponding to the distribution of the parameters $\wV^{b}$ given all the other variables.}
	\label{fig:chain_gtr_factor_graph}
\end{figure}

Let us illustrate this on the chain \GTR\ model introduced in Section~\ref{sec:model}.
Specifically, we will next present a concrete example where the \LBPS-\HMC\ alternation strategy achieves strong sparsity. 
We show in Figure~\ref{fig:chain_gtr_factor_graph} the factor graph corresponding to the distribution of the parameters $\wV^{b}$ given all the other variables. 

Regardless of the size of the state space $\statespace$, we next argue that for any $\{x, x'\}\in \statespaceup$, any sojourn time factor $H_{x, x'}$ or transition count $C_{x, x'}$ has at most twelve extended neighbour factors and three extended neighbour variables under Bayesian \GLM\ chain \GTR\ when conditioning on $\wV^{u}$. For the pairs of states $\{x, x'\}$, when $\eta(\{x, x'\})\ne\vert\statespace\vert(\vert\statespace\vert-1)/2$, the twelve extended neighbour factors of factor $H_{x, x'}$ or $C_{x, x'}$ include 
the factors only connected with $\wV^b_i$ and $\wV^b_{i+1}$, where $i=\eta(\{x, x'\})-2, \eta(\{x, x'\})-1,$ and $\eta(\{x, x'\})$. Similarly, for the pair of states $\{x, x'\}$ such that $\eta(\{x, x'\})=\vert\statespace\vert(\vert\statespace\vert-1)/2$, $H_{x, x'}$ or $C_{x, x'}$ has eight extended neighbour factors, where each of the eight factors is only connected with $\wV^b_{i}$ and $\wV^b_{i+1}$, where $i=\eta(\{x, x'\})-2$ and $\eta(\{x, x'\})-1.$ All factors have three extended neighbour variables $\wV^b_{\eta(\{x, x'\})-1}, \wV^b_{\eta(\{x, x'\})}$ and $\wV^b_{\eta(\{x, x'\})+1}$, except that factor $H_{x'', x'''}$ or $C_{x'', x'''}$ has two extended neighbour variables $\wV^b_1$ and $\wV^b_2$ when $\eta(\{x'', x'''\})=1$.

Thus, when updating the position of extended neighbour variables or computing the candidate bounce time for extended neighbour factors of the collision factor, there is only a small constant number of extended neighbour variables and extended neighbour factors involved.  We can conclude that the factor graph in Figure~\ref{fig:chain_gtr_factor_graph} satisfies the desirable strong sparsity in definition~\ref{def:strong_sparsity}.

\subsection{Simulation of bounce inter-arrival times for \GLM\ \GTR\ models}
\label{sec:solution_time}

In this section, we show how to simulate the candidate times $\tau_f$ for the next bounce for each factor $f$. 
The simulation algorithm for a Gaussian distribution can be found in \citet{bouchard2018bouncy}, so we focus on the other factors.

\subsubsection{Sojourn time factors}
\label{sec:sojourn_time_solution}

We first sample the energy gap $-\log(E)>0$ with $E\sim\mathcal{U}(0,1)$. This gap represents the difference between the energy denoted as $E_0$ at the current position of a particle and a higher energy $E_0-\log(E)$. The goal is to determine a time interval $\delta$ such that at this time point the particle has an energy of $E_0-\log(E)$.
 
For sojourn factors, the potential energy of the particle is:


\begin{eqnarray*}
U(\wV_0)&=& \h^{(\mcmciter)}_\x  \qrev^\rev_{\x, \x'}(\wV_0)\\
&=& \h^{(\mcmciter)}_\x \pi_{x'}\exp(\dotprod{\wV_0}{ \sufftransV(\{\x,\x'\}) }).
 \end{eqnarray*}
The potential energy after a time interval $\delta$:
\begin{eqnarray*}
U(\wV_0+\boldsymbol{v}\delta)&=& \h^{(\mcmciter)}_\x  \qrev^\rev_{\x, \x'}(\wV_0+\boldsymbol{v}\delta)\\
&=& \h^{(\mcmciter)}_\x \pi_{x'}\exp( \dotprod{\wV_0+\boldsymbol{v}\delta}{\sufftransV(\{\x,\x'\})} ).
 \end{eqnarray*}
 We observe that the particle is travelling to a higher energy area if and only if $\dotprod{\boldsymbol{v}}{\sufftransV(\{\x,\x'\})}>0$.
Therefore, we set
 \begin{eqnarray*}
-\log(E) &=& U(\wV_0+\boldsymbol{v}\delta)- U(\wV_0)\\
&=& \h^{(\mcmciter)}_\x  \qrev^\rev_{\x, \x'}(\wV_0)\bigg(    \exp(\dotprod{\boldsymbol{v}\Delta}{ \sufftransV(\{\x,\x'\}) }) -1\bigg).
\end{eqnarray*}

We denote $c=-\log(E)>0$ and obtain: 

\[
    \delta = \begin{cases}
       \frac{1}{\dotprod{\boldsymbol{v}}{\sufftransV(\{\x,\x'\})}}\log\left(
 \frac{c}{\h^{(\mcmciter)}_\x\qrev^\rev_{\x, \x'}(\wV_0)} +1\right) & \text{if} \quad \dotprod{\boldsymbol{v}}{\sufftransV(\{\x,\x'\})} >0,\\
        \infty, & \text{otherwise.} 
        \end{cases}
  \]

 \subsubsection{Transition count factors}
 \label{sec:transition_count_solution}
 Similarly, the transition count factor for pairs of states $(x, x')$ is $-\c^{(\mcmciter)}_{\x, \x'} \log\left( \qrev^\rev_{\x, \x'}(\wV_0)\right)$ in the $i$th iteration, and the gradient is $-\c^{(\mcmciter)}_{\x, \x'}\sufftransV(\{\x,\x'\})$. Thus, the  potential energy of the particle is:
\begin{eqnarray*}\label{eq:gradient_of_transition_count_gtr}
U(\wV_0) &=& -\c^{(\mcmciter)}_{\x, \x'} \log\left( \qrev^\rev_{\x, \x'}(\wV_0)\right)\nonumber\\
&=&  -\c^{(\mcmciter)}_{\x, \x'} \bigg(
\log\pi_{x'}+\dotprod{\wV_0}{\sufftransV(\{x, x'\})}
\bigg)
\end{eqnarray*}
The potential energy after a time interval $\Delta$: 
\begin{eqnarray*}
U(\wV_0+\boldsymbol{v}\delta) &=& -\c^{(\mcmciter)}_{\x, \x'}\bigg( 
\log(\pi_{x'})+\dotprod{\wV_0+\boldsymbol{v}\delta}{\sufftransV(\{x, x'\})}
\bigg)
\end{eqnarray*}
We sample $E\sim\mathcal{U}(0, 1)$, set $c=-\log(E)$, $U(\wV_0+\boldsymbol{v}\delta) - U(\wV_0) =c$, and obtain: 

\[
    \delta = \begin{cases}
     -\frac{c}{\c^{(\mcmciter)}_{\x, \x'}\dotprod{\boldsymbol{v}}{\sufftransV(\{x,x'\})}} & \text{if} \quad \dotprod{\boldsymbol{v}}{\sufftransV(\{\x,\x'\})} \leqslant 0,\\
        \infty, & \text{otherwise.}
	\end{cases}
\]
\subsection{Running time analysis of \LBPS-\HMC}\label{sec:running_time_chain_gtr}
In Section~\ref{sec:running_time_lbps_general}, we have provided the running time analysis of \LBPS\ under a general factor graph $G$. In this section, we analyze the running time 
of \LBPS-\HMC\ under the Bayesian \GLM\ chain \GTR\ model for \CTMCs\ by simplifying the computational cost of a factor graph $G$ with sparse property under the condition that 
$c_{*}=\mathcal{O}(1)$ and $c_{U_{*}}=\mathcal{O}(1)$, which is satisfied in the Bayesian \GLM\ chain \GTR\ model. The condition that $c_{*}=\mathcal{O}(1)$ and $c_{U_{*}}=\mathcal{O}(1)$ indicates that the computational cost of computing the collision time and the gradient in terms of $\wV$ for any factor in the factor graph is of constant time regardless of the dimension of the state space $\mathcal{X}$ in the \CTMC. Recall that the computational cost for a strongly sparse factor graph $G$ is given in Equation~\ref{eq:compute_cost_sparsity}.

The computational cost we derive not only holds for Bayesian \GLM\ chain \GTR\ model but also holds for any Bayesian \GLM\ rate matrix model that satisfies $c_{*}=\mathcal{O}(1), c_{U_{*}}=\mathcal{O}(1),  \left\vert\overline{N}_*\right\vert=\mathcal{O}(1)$ and $ \left\vert\overline{S}_*\right\vert=\mathcal{O}(1)$. Under the Bayesian \GLM\ chain \GTR\ rate matrix parameterization, we have $m=\mathcal{O}\left(\vert\statespace\vert^2\right)$ and $p=\mathcal{O}\left(\vert\statespace\vert^2\right)$. 

The running time of the key steps in Algorithm~\ref{alg:algo_lbps} is as follows:

\begin{enumerate}
\item Sample the auxiliary sufficient statistics using the end-point sampler: $\mathcal{O}\left(\vert\statespace\vert^3\right)$.
\item Update the univariate weights $\wV^u$ using \HMC: $\mathcal{O}(J_0\vert\statespace\vert)$, where $J_0$ represents the number of leapfrog steps in one \HMC\ iteration. When updating $\wV^u$, the number of non-zero entries for all possible features is $\vert\statespace\vert$. 
\item Update the bivariate weights $\wV^b$ using \LBPS: $\mathcal{O}( J_1\log\vert\statespace\vert+
J_2\log\vert\statespace\vert ).$
This is obtained by plugging $\alpha_1=\alpha_2=0$ into Equation~\ref{eq:compute_cost_sparsity}, where $\alpha_1=\alpha_2=0$ since we have $\left\vert\overline{N}_*\right\vert=\mathcal{O}(1)$ and $ \left\vert\overline{S}_*\right\vert=\mathcal{O}(1)$. A local refreshment scheme is adopted. 
\end{enumerate}

Thus, the total cost of one iteration of our \LBPS-\HMC\ alternation algorithm is 
\begin{eqnarray}
\mathcal{O}(J_0\vert\statespace\vert+J_1\log\vert\statespace\vert+
J_2\log\vert\statespace\vert +\vert\statespace\vert^3).
\label{eq:LBPS_cost}
\end{eqnarray}
In comparison, one \HMC\ iteration for Bayesian \GLM\ chain \GTR\ takes $\mathcal{O}\left( J_0\vert\statespace\vert^2+\vert\statespace\vert^3\right)$ since the total number of non-zero entries for all possible features of Bayesian \GLM\ chain \GTR\ is $\vert\statespace\vert(\vert\statespace\vert+1)/2$ using only \HMC, where $J_0$ is the number of leapfrog steps in one \HMC\ iteration. In our experiments, we find that $\vert\statespace\vert$, $\log\vert\statespace\vert$ are both lower order terms compared with $\vert\statespace\vert^2$ using \HMC. This potentially explains why our \LBPS-\HMC\ is more computationally efficient than \HMC. The running time analysis discussed here is not a whole story since we do not provide information on how $J_1$ and $J_2$ should be scaled in order to obtain a constant number of effective samples. To provide the readers an idea of how $J_0$ in \HMC\ scales, consider the case of normally distributed random vectors of dimension $p$ with an identity covariance matrix.  To reach a nearly independent point, the number of leapfrog updates under the independent and identically distributed case grows as $\mathcal{O}\left(p^{\frac{1}{4}}\right)$ \citep{neal2011mcmc}, where $p$ is the dimension of the parameters. For \LBPS, under the independent and identically distributed normal distribution with dimension $p$ or weakly dependent case, we expect $J_1$ to grow as $O(p)$.

\section{Experiments}\label{sec:experiments}
All numerical experiments are run on Compute 
Canada resources ``cedar" with Intel ``Broadwell" CPUs at 2.1Ghz and an E5-2683 v4 model. A detailed description of ``cedar" can be found at \url{https://docs.computecanada.ca/wiki/Cedar}. The algorithm is implemented in Python 3.6.4 and available at \url{https://github.com/zhaottcrystal/rejfreePy_main}.
Across all experiments, the \ESS\ is evaluated via the R package \emph{mcmcse}~\citep{flegal2012mcmcse}. We set the first 30\% of the posterior samples as the burnin period to be discarded.

All the synthetic datasets are simulated under a Bayesian \GLM\ chain \GTR\ rate matrix parameterization. We assign a standard Gaussian distribution as the prior over both the univariate and bivariate weights, which are both generated from $\mathcal{U}(0, 1)$ across all simulation studies. The refreshment rate of \LBPS\ is set to one and the number of leapfrog jumps and the stepsize of each jump are set as $L=40$ and $\epsilon=0.001$ in \HMC. The combination of the two tuning parameters are simply chosen among the ones with the best computational efficiency given multiple combination of the parameter values. 

Across all synthetic experiments, we generate 500 sequences given a total length of the observation time interval as three and the states of all sequences are observed every 0.5 unit time under a Bayesian \GLM\ chain \GTR\ with different dimensions of rate matrices. Later in this paper, we use ``LBPS" for short to represent the combined sampling scheme \LBPS-\HMC, which uses \HMC\ to sample the univariate weights and \LBPS\ to sample the bivariate weights. When \HMC\ sampling scheme is mentioned later, it refers to the sampling scheme using \HMC\ to sample all the weight parameters $\wV=\left(\wV^u, \wV^b\right)$. 

In both synthetic data experiments in Section~\ref{sec:experiments} and real data analysis in  
Section~\ref{sec:real}, we conduct computational efficiency comparison via \ESS\ per second and we also check the correctness of the two sampler implementation. For the synthetic data experiments, we perform \EIT\ \citep{bouchard2019blang} in the spirit of \citet{geweke2004getting}. We compare the density of the posterior samples between \LBPS\ and \HMC, and evaluate their difference using \ARD\ (defined in Appendix I). Since both algorithms share the same limiting distribution, as we increase the number of iterations of the Markov chains, after an initial burnin period, the distributions of the posterior samples from the two algorithms are expected to be closer to the target posterior distribution. For higher dimensional rate matrices, since the total number of parameters is $\mathcal{O}(\vert\statespace\vert^2)$, it is hard to display the density plots for all parameters, we use \ARD\ as the metric to describe the similarities between posterior samples from the two algorithms. Smaller values of \ARD\ indicate more similarities. 
In Section~\ref{sec:real} with the real dataset, we also compute the \ARD\ and report the results in Appendix I.

\subsection{Correctness check of \LBPS\ and \HMC\ using EIT}

To check the correctness of our software implementation,
we perform out testing procedure \EIT, a simple extension of \citet{geweke2004getting}, with details provided in Appendix D. We performed tests on both \LBPS\ and \HMC\ kernels on various dimensions of the rate matrices (see Appendix D). 

\begin{figure}[ht]
  \begin{center}
   \includegraphics[scale=0.5]
    {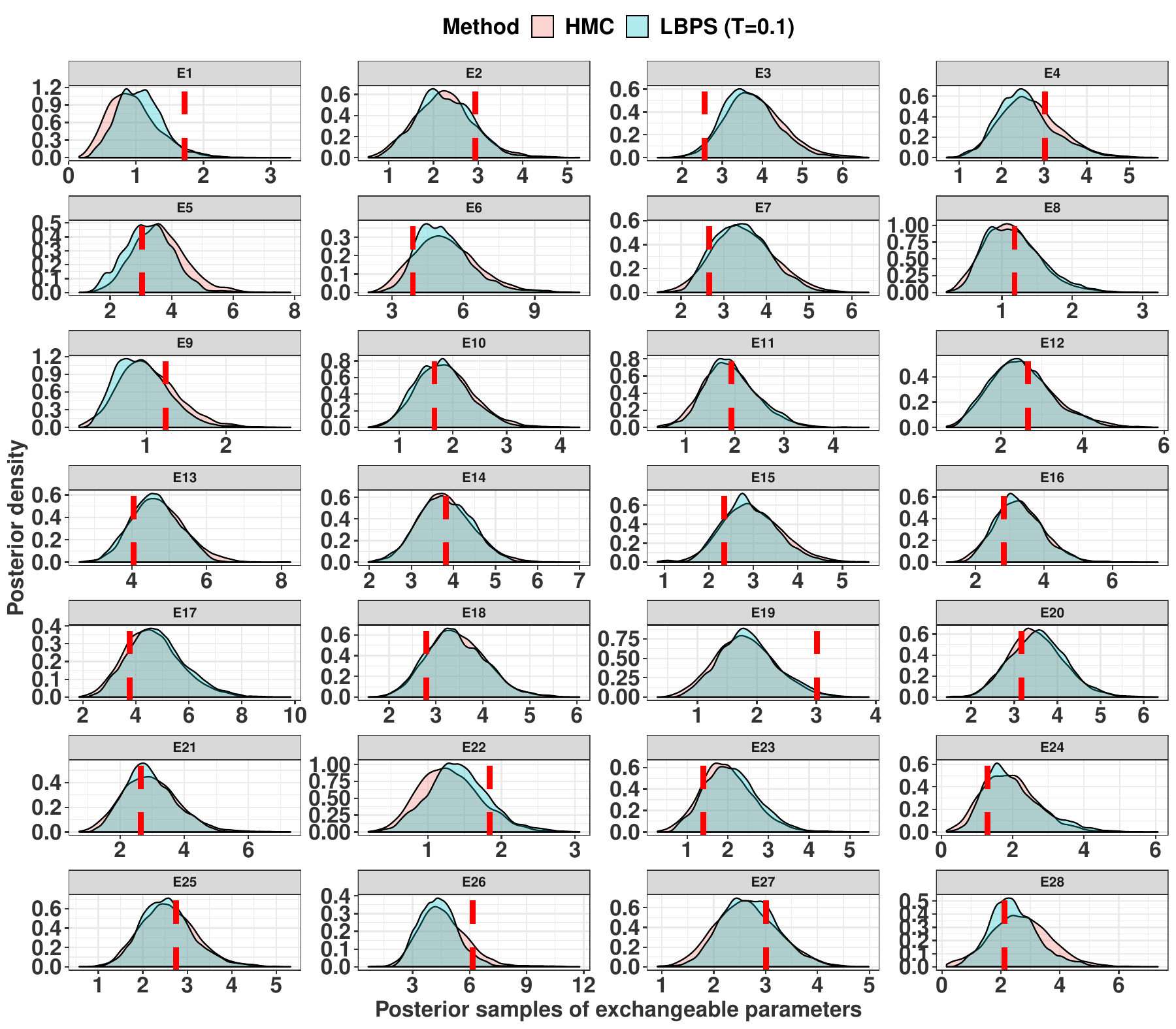}
  \end{center}
\caption{ Density plot of posterior density comparison across exchangeable parameters of an 8-by-8 rate matrix between \LBPS\ (with trajectory length 0.1) and \HMC.}
\label{fig:densityplot}
\end{figure}

To further validate the correctness of \LBPS\ and \HMC, we compare the distribution of the posterior samples collected from \LBPS\ and \HMC\ respectively. The shared synthetic dataset is generated under an 8-by-8 Bayesian \GLM\ chain \GTR. The prior distribution for $\wV$ is standard Gaussian distribution. We obtain 40,000 and 10,000 posterior samples from \LBPS\ and \HMC\ separately with first 30\% of the samples discarded. With a long chain from both algorithms, we expect their density plots to be close and we demonstrate this  in Figure~\ref{fig:densityplot} using the density plot of all exchangeable parameters. The boxplot is provided in Figure~\ref{fig:boxplot} in Appendix E. We also provide summary statistics of the posterior samples among different parameters. 

\subsection{Computational efficiency comparison between \LBPS\ and \HMC}
 
In order to explore the scalability of our algorithm as the dimension of the parameters increases, we compared the \ESS\ per second in the $\log_{10}$ scale among all the parameters for different sizes of the rate matrices. Larger \ESS\ per second indicates better computational efficiency of the algorithm. One obvious choice to compare the efficiency of MCMC algorithms is the average MCMC efficiency over all parameters, which is represented by the mean \ESS\ per second. However, the algorithm efficiency is better represented by the minimum ESS per second since the entire posterior samples are valid unless all parameters have mixed adequately. Thus, we  focus on the minimum ESS per second over all parameters.

\begin{figure}[t]
  \begin{center}
   \includegraphics[scale=0.55]
    {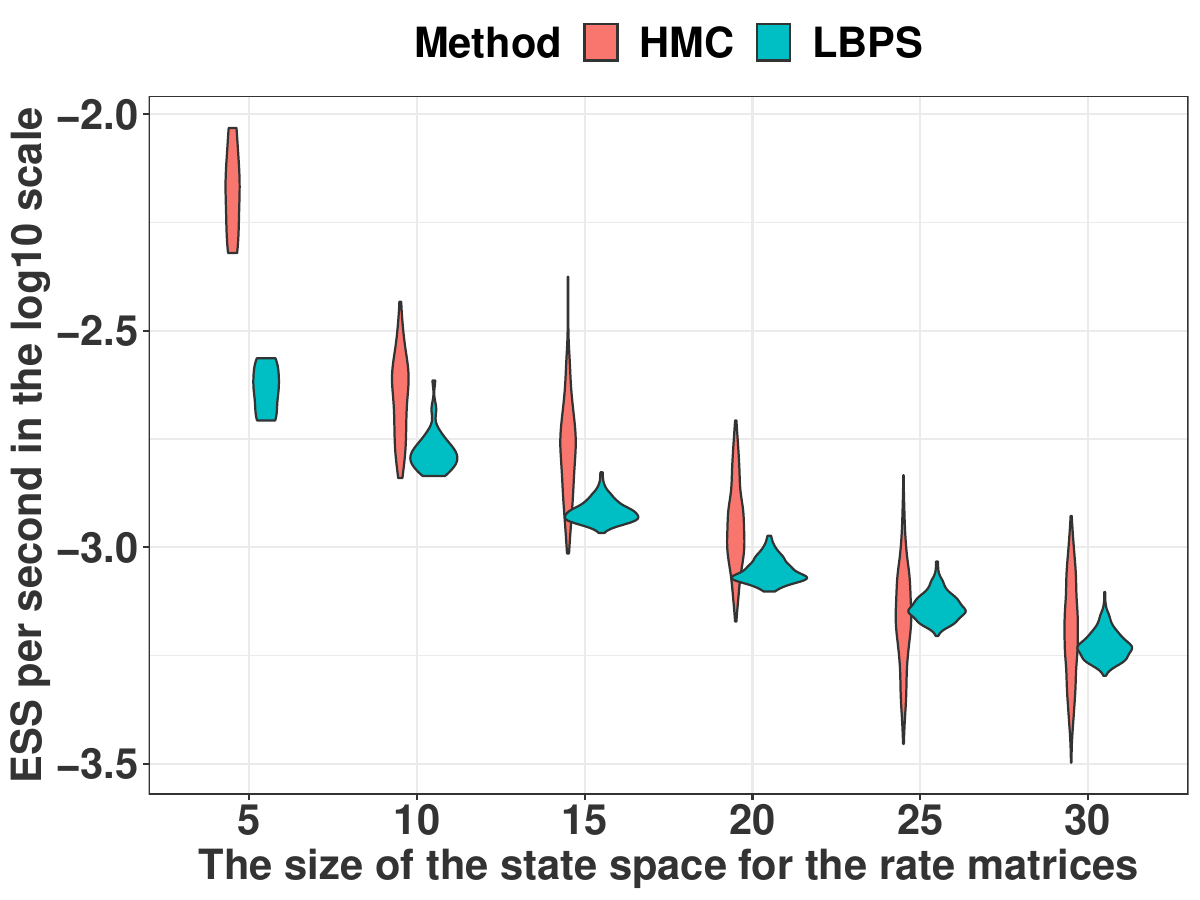}
  \end{center}
\caption{ \ESS\ per second in the $\log_{10}$ with the dimension of the rate matrices ranging from 5-by-5, 10-by-10, $\ldots$ to 30-by-30 with a step size of 5.}
\label{fig:ess_per_second}
\end{figure}

The dimension of the rate matrices ranges from 5-by-5, 10-by-10, $\ldots$ to 30-by-30. We obtain a total number of 60,000 posterior samples from \LBPS\ and 10,000 samples from \HMC. To speed up the actual running time of the experiments, for rate matrices of dimension lower than 15-by-15, the trajectory length is fixed as 0.1. For rate matrices with higher dimensions, the trajectory length is shortened to 0.05. Under the Bayesian \GLM\ chain \GTR\ model parameterization, the number of exchangeable parameters is $\vert\statespace\vert(\vert\statespace\vert-1)/2$. The result is displayed in Figure~\ref{fig:ess_per_second}. 

From Figure~\ref{fig:ess_per_second}, we can see that when the dimension of the weight parameters is low, for a 5-by-5 rate matrix, \HMC\ has better performance than \LBPS. As the dimension increases, it shows that the minimum of the ESS per second for \LBPS\ outperforms \HMC. The larger variation in \ESS\ per second of \HMC\ in the violin plot in Figure~\ref{fig:ess_per_second} indicates that \HMC\ is not efficient in exploring certain directions of the parameter space. We find that the minimum ESS per second and its lower quantiles for LBPS are better than HMC as the dimension increases. This result suggests that LBPS has better computational efficiency than HMC for the class of high-dimensional CTMCs considered in this section.



\subsection{\ESS\ per second using different trajectory lengths of \LBPS\ comparisons}\label{sec:trajectoryLength}
The computation efficiency of \HMC\
has been found to depend highly on the choices of the tuning parameters, which are the number of leapfrog jumps $L$ and the size of each jump $\epsilon$. Various strategies~\citep{wang2013adaptive, hoffman2014no, zhao2016bayesian} have been developed to effectively tune the parameters. \LBPS\ also involves a tuning parameter, which is the fixed trajectory length. Thus, it is of interest to examine whether the computational efficiency of \LBPS\ is sensitive to different choices of the trajectory length. 

\begin{figure}[t]
  \begin{center}
   \includegraphics[scale=0.3]
   {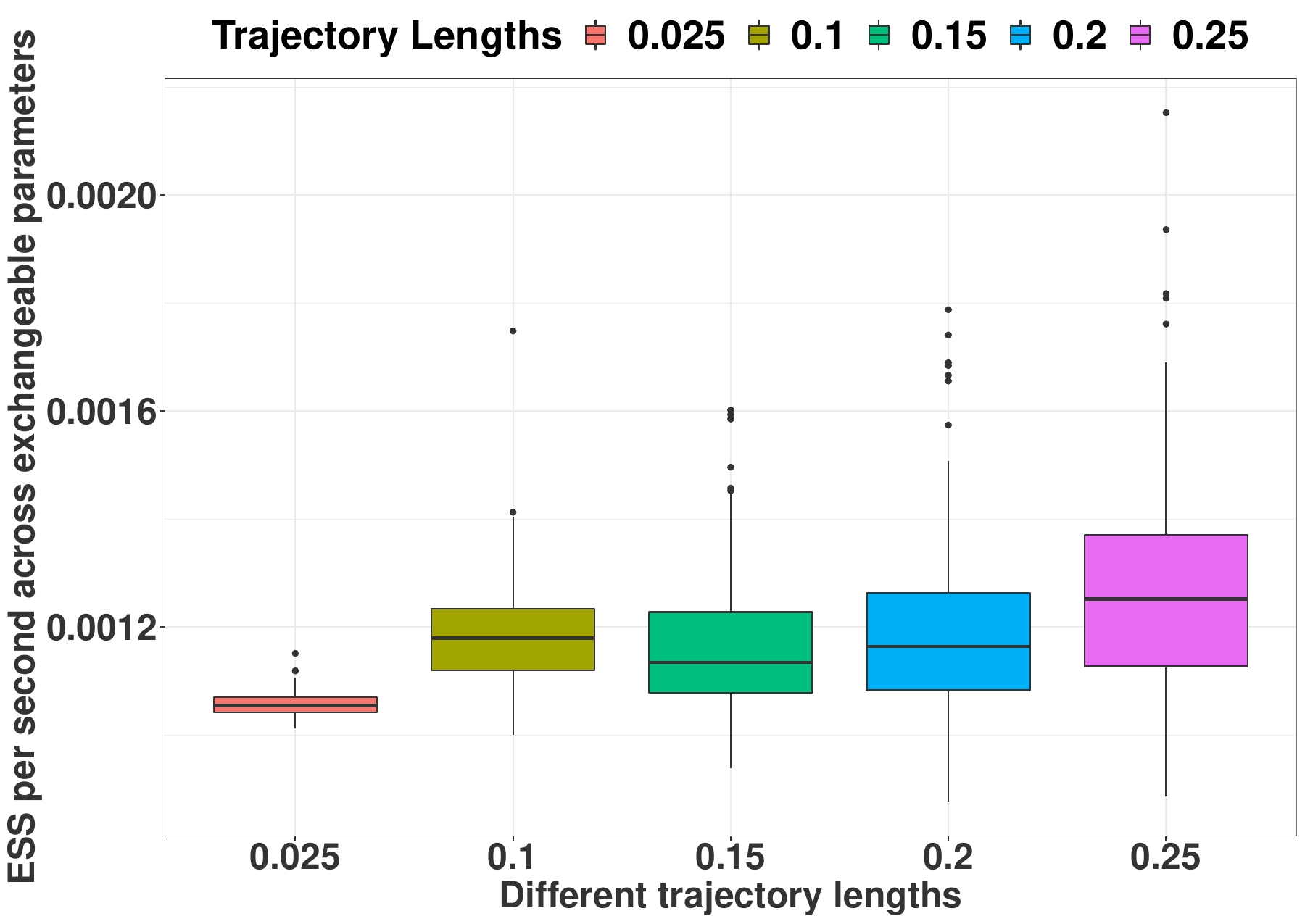}
  \end{center}
\caption{\ESS\ per second of the summary statistics across exchangeable parameters with respect to different trajectory lengths of \LBPS.}
\label{fig:ess_per_second_traj}
\end{figure}

We simulate a synthetic dataset from a 20-by-20 rate matrix with 190 exchangeable parameters. A 20-by-20 rate matrix is chosen since it has the same dimension as protein evolution in the real data analysis. We use \ESS\ per second as the metric to evaluate the computation efficiency of \LBPS. Our summary statistics include the minimum, first quantile, mean, median, third quantile and maximum of the \ESS\ per second across 190 exchangeable parameters. We obtain 40,000 posterior samples of the parameters of interest.  We show the actual walltime (in seconds) of our algorithm with fixed trajectory lengths at 0.025, 0.1, 0.15, 0.2 and 0.25  in Table~\ref{table:walltime}. Since the first 30\% of the samples are discarded, the actual running time used to compute the \ESS\ per second are scaled by 70\% of the total walltime in Table~\ref{table:walltime}.
\begin{table}[H]
  \begin{center}
   \begin{tabular}{cccccc}
   \toprule
     Trajectory Lengths& 0.025 & 0.10 & 0.15 & 0.20 & 0.25\\
     \midrule
     Walltime (in seconds) & 240677&282233&338242&381565&403932\\
     \bottomrule
    \end{tabular}
   \end{center}
 \caption{Actual walltime of \LBPS\ for 40,000 iterations with trajectory lengths at 0.025, 0.1, 0.15, 0.2, 0.25.}
 \label{table:walltime}
\end{table}
The results are shown in Figure~\ref{fig:ess_per_second_traj}. We have found that except the maximum of the \ESS\ per second, the computational efficiency of \LBPS\ is similar across different trajectory lengths and especially the minimum of \ESS\ per second is very robust. The longer the trajectory lengths, the better performance of the maximum of the \ESS\ per second. Similar conclusions are achieved in our real data analysis. The minimum of the \ESS\ per second is more important than the maximum since the performance of a sampler depends on the direction that is the hardest to sample.  

\section{Real Data Analysis}\label{sec:real}
\subsection{Background}
We use the real dataset from \citet{zhao2016bayesian} with 641 amino acid sequences and each sequence has 415 sites from the protein kinase domain family. It is available at \url{https://github.com/zhaottcrystal/rejfreePy_main/tree/master/Dataset}. In phylogenetics, the evolutionary process is often inferred using multiple homologous biological sequences under a evolutionary tree with the same rate matrix across the tree. For simplicity, we 
estimate the rate matrix from a pair of sequences. We pick randomly a pair of amino acid sequences to study the rate matrix from its posterior distribution.  

In Section~\ref{sec:Bayesian GLM}, we have provided intuition behind the chain \GTR\ model since we assume that the exchangeable rates 
between a pair of states is affected by its neighbour pairs with similar biological properties. Then another question arises as how to pick a reasonable ordering of pairs of amino acids to allow neighbour pairs share similar biological properties. According to \citet{he2011amino}, we determine the ordering according to the closeness among amino acids based on their pairwise Euclidean distance defined by their physiochemical properties. The distance satisfies nonnegative, reflective, symmetric, and triangle properties and is given by \citet{grantham1974amino} as :
\begin{eqnarray}
\label{eq:distance}
D_{ij} = \bigg(\alpha(c_i-c_j)^2 + \beta(p_i-p_j)^2 + \gamma(v_i-v_j)^2   \bigg)^{\frac{1}{2}},
\end{eqnarray}
where $c=\textrm{composition}$, $p=\textrm{polarity}$, $v=\textrm{molecular volume}$, and $\alpha, \beta,\gamma$ are defined in \citet{grantham1974amino}. Larger pairwise distance indicates less similarity between pairs of amino acids, where mutation between them may be deteriorative. On the contrary, smaller distance indicates more similarities and easier mutations. The pairwise Euclidian distance between amino acids is given in Table 3 by \citet{he2011amino}. We provide the table in Appendix F.  

Given the distance in Appendix F, the ordering of pairs of amino acids regarding their closeness is determined according to our proposed \NNPAAO\ Algorithm~\ref{alg:ordering} in Appendix G. Some notations are introduced to help understand the algorithm. Denote $\textrm{AminoAcidDist}$ as the pairwise Euclidean distance between amino acids shown in Appendix F, $\statespace$ as the state space of 20 amino acids and $\statespaceup$ as the set of unordered pairs of distinct amino acids, where $|\chi|=20$ and $\left\vert\statespaceup\right\vert = 190$. The algorithm outputs a dictionary $\textrm{AminoAcidsPairRank}$ with keys from $\statespaceup$ and the value associated with each key represents the rank of this amino acid pair. Neighbour pairs indicate closeness for similarities.

The algorithm first picks the amino acid pair with the smallest positive Euclidean distance, which is pair ``IL", where D(I, L) = 5.  The nearest neighbour to ``IL" is chosen among two candidate pairs with the smallest nonzero distance to ``I" or ``L", which are ``IM" with D(I, M)=10 or ``LM" with D(L, M)=15. Among them, we pick the pair with a smaller distance and ``IM" is defined as the nearest neighbour to ``IL". Next our current pair is  set as ``IM" and similarly, we find the nearest neighbour to ``IM" and we keep searching the nearest neighbour for the current pair until we have iterated over all pairs in $\statespaceup$.

\subsection{Numerical results: computational efficiency comparison}\label{sec:computational_efficiency_comparison}
\begin{figure}[H]
  \begin{center}
   \includegraphics[scale=0.2]
   {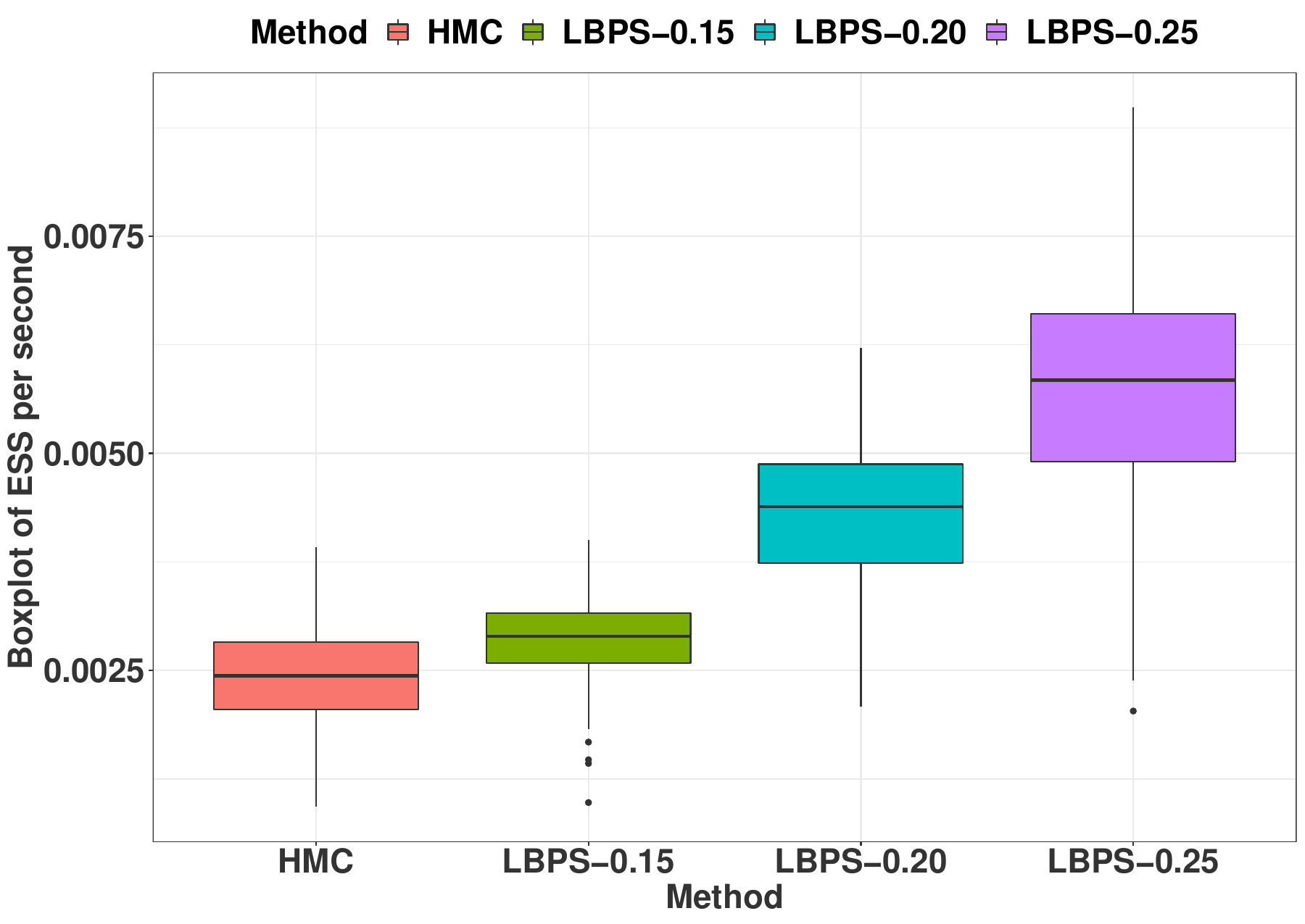}
  \end{center}
\caption{Summary statistics of \ESS\ per second of across exchangeable parameters with different trajectory lengths of \LBPS\ at 0.1, 0.15 and 0.2, compared to \HMC.}
\label{fig:ess_per_second_real_data}
\end{figure}

We provide the correctness check via \ARD\ between the two samplers in Appendix~I. We compare the computational efficiency of \LBPS\ and \HMC\ via comparing the summary statistics of the \ESS\ per second across 190 exchangeable parameters of a 20-by-20 reversible rate matrix using the protein kinase domain family dataset in \citet{zhao2016bayesian}. Figure~\ref{fig:ess_per_second_real_data} demonstrates the computational advantages of using \LBPS\ over \HMC. The trajectory lengths of \LBPS\ is chosen as 0.1, 0.15 and 0.2. We find that when the trajectory length is set at 0.2, \LBPS\ outperforms \HMC\ across all summary statistics of \ESS\ per second. The minimum of \ESS\ per second of \HMC\ is markedly worse than \LBPS, this implies that the inefficiency of \HMC\ to explore certain direction of the parameter space. 
Under a fixed trajectory length of \LBPS, there is no big difference between the first quantile and third quantile of the \ESS\ per second across all parameters. It indicates that \LBPS\ has similar computational efficiency across different directions of the parameter space. In Table~\ref{tb:ess_lbps_hmc_real_data}, we provide the \ESS\ runing \HMC\ for 811380 seconds  (9.4 days) compared to \LBPS\ running  only 81.25\% of the walltime of \HMC, which further demonstrates the superior performance of \LBPS\ compared to \HMC. A traceplot scaled by running time of a selected exchangeable parameter showing better mixing using \LBPS\ compared to \HMC\ is provided in Figure~\ref{fig:traceplot_real_data} in Appendix H. 

 \begin{table}[ht]
  \begin{center}
 \begin{tabular}{lcccccc}
 \toprule
 &\multicolumn{6}{c}{Summary Statistics} \\
\cmidrule(lrrrrrr){2-7}
 Method  &    Min. & 1st Quantile& Median & Mean & 3rd Quantile & Max.\\
    \midrule
  \LBPS\ (0.2) &   773  & 1656   & 1864   & 1845 &   2108   & 2532 \\
  \midrule
  \HMC\ &  95  &   626 &    740 &    736 &     840 &   1233 
    \\
     \bottomrule
    \end{tabular}
   \end{center}
 \caption{Summary of \ESS\ across exchangeable parameters between \HMC\ and \LBPS\ with trajectory length 0.2.}
 \label{tb:ess_lbps_hmc_real_data}
\end{table}

\section{Discussion}\label{sec:extensions}
In this paper, we have developed a computationally efficient sampling scheme combining \LBPS\ and \HMC\ to explore high dimensional \CTMCs\ under a novel Bayesian \GLM\ chain \GTR\ rate matrix. In this model, we assume that the mutation rates of the amino acid evolutionary processes depend on its neighbour pairs with similar physiochemical properties. 

We also provide a framework for assessing the running time of \LBPS\ algorithm, based on a notion of \emph{strong sparsity}. 
In terms of empirical performance, we found that in the problems considered, \HMC\ alone leads to inefficient mixing of certain directions of the parameter space. In the contrary, the computational performance of our method is more uniform across different directions of the parameter space. In the real application, our method outperforms stand-alone \HMC\ in terms of \ESS\ per second for all chosen summary statistics.

Moreover, we provide the first proof-of-concept real data application applying \LBPS\ to  high dimensional \CTMCs, showing the great potential of this computational method in this domain. This shows promises for evolutionary processes  with a large state space, such as codon evolution~\citep{anisimova2008investigating}. Our sampling algorithm also has the potential to be applied to co-evolution of groups of interacting amino acids residues which also involves large rate matrices. 

For phylogenetic applications, one of the limitations of our algorithm in its current form is that it assumes an unnormalized reversible rate matrices since we were only able to obtain  analytical solution to the collision time under this situation. It is customary in phylogenetics to use normalized rate matrix, where the normalization ensures that the expected number of mutations is one given one unit time. To resolve this issue, one can normalize the branch lengths instead of the rate matrix to recover identifiability. A related issue is that changing the prior on the weights requires that the algorithms used to simulate the collision times need be adapted. Fortunately several methods can be used to approach these collision time simulation algorithms, see e.g.\ \cite{bouchard2018bouncy}, Section 2.3.  

As another direction for future work, it would be interesting to combine our method with Bayesian variable selection techniques \citep{ishwaran2005spike} and a binary version of \BPS\ \citep{pakman2017binary} to select subset of features that have a large effect on the underlying \CTMCs.






\newpage
\appendix
\section*{Appendix A. Connection between Bayesian \GLM\ \GTR\ model and Bayesian \GLM\ chain \GTR\ model}
We have defined $\sufftransV^{\textrm{gtr}}$ in Section~\ref{sec:Bayesian GLM} such that any \GTR\ rate matrices can be represented via the Bayesian \GLM\ rate matrix parameterization. Our objective is to prove that there exists $\wV_{*}$ such that for any rate matrices under the Bayesian \GLM\ parameterization, for $\forall \wV, \qrev^\rev_{\x, \x'}(\wV)$ can be represented under the Bayesian \GLM\ chain \GTR\ parameterization. 

\citet{zhao2016bayesian} have shown that for any rate matrix $Q$, we can find weights $\wV$ such that $Q_{(x, x')}=\qrev_{\x, \x'}(\wV)=\exp\big\{\dotprod{\wV^b}{\sufftransV^{\textrm{gtr}}(\{\x,\x'\})}\big\}\pi_{\x'}(\wV^u)$ under the Bayesian \GLM\ \GTR\ model. In the Bayesian \GLM\ chain \GTR\ model, we show that there exists $\wV_*$ such that for $\forall \wV=\begin{pmatrix}
\wV^u\\
\wV^b
\end{pmatrix}$, 

\begin{eqnarray*}
\qrev^{\textrm{gtr}}_{\x, \x'}(\wV)&=&\exp\big\{\dotprod{\wV^b}{\sufftransV^{\textrm{gtr}}(\{\x,\x'\})}\big\}\pi_{\x'}(\wV^u)\\
&=&\qrev^{\textrm{chain}}_{\x, \x'}(\wV_*)\\
&=&\exp\big\{\dotprod{\wV^b_*}{\sufftransV^{\textrm{chain}}(\{\x,\x'\})}\big\}\pi_{\x'}(\wV^u_*)
\end{eqnarray*}
Thus, we can set $\wV_*^{u}=\alpha\wV^{u}$ for any $\alpha \in \mathbb{R}$ so that $\pi_{\x'}(\wV^u)=\pi_{\x'}(\wV^u_*),$ for $\forall x'\in\statespace$. By plugging the definition of $\sufftransV^{\textrm{gtr}}$ and $\sufftransV^{\textrm{chain}}$ in Equation~\ref{eq:gtr_phi} and Equation~\ref{eq:chain_phi} respectively, we require that:
$$
\begin{cases}
\left(\wV_*^b\right)_{\eta(\{x, x'\})}+\left(\wV_*^b\right)_{\eta(\{x, x'\})-1}=\left(\wV^b\right)_{\eta(\{x, x'\})}, &\text{for}\,\, \eta(\{x, x'\})=2, 3, \ldots, \statespace|(|\statespace|-1)/2),\\
\left(\wV_*^b\right)_1=\left(\wV^b\right)_1, &\text{for}\,\, \eta(\{x, x'\})=1.
\end{cases}
$$

Set $p=\statespace|(|\statespace|-1)/2)$, it is equivalent to solve 
\begin{align*}
\left(\wV_*^b\right)_1&=\left(\wV^b\right)_1\\
\left(\wV_*^b\right)_1+\left(\wV_*^b\right)_2 &= \left(\wV^b\right)_2\\
\left(\wV_*^b\right)_2+\left(\wV_*^b\right)_3 &= \left(\wV^b\right)_3\\
&\vdots\\
\left(\wV_*^b\right)_{p-1}+\left(\wV_*^b\right)_{p}&= \left(\wV^b\right)_p.
\end{align*}
We obtain the solution
$\wV^{b}_*=\boldsymbol{B}\wV^b$, where $$
\boldsymbol{B}_{ij}=
\begin{cases}
(-1)^{i+j-2}, &\text{if}\,\,  j\leqslant i\,\text{and}\, \, i, j=1, 2, \ldots, (|\statespace|(|\statespace|-1)/2), \\
0, &\text{otherwise}.
\end{cases}
$$
Thus, the solution exists, where $\wV_*=\begin{pmatrix}
\wV^u_*\\
\wV^b_*
\end{pmatrix}
=\begin{pmatrix}
\alpha\wV^u\\
\boldsymbol{B}\wV^b
\end{pmatrix}.$

%

\section*{Appendix B. Gradient Computation for Univariate Weights in the \HMC\ step under combined sampling scheme}
\label{app:uni_theorem}

In our proposed sampling scheme \LBPS-HMC\ described in Algorithm~\ref{alg:algo_lbps}, we use \HMC\ to update only the univariate weights $\wV^u$ while fixing the bivariate weights $\wV^{b}$ and the auxiliary variable $\boldsymbol{z}$ and \LBPS\ to update the bivariate weights $\wV^{b}$ while fixing $\wV^u$ and  $\boldsymbol{z}$. We provide the gradient information for $\wV^u$ required by \HMC\ of our combined sampling scheme under a reversible, unnormalized rate matrix.

We review that the augmented joint density given the sufficient statistics $\zV(\yV) \defeq (\nV(\yV), \hV(\yV), \cV(\yV))$ for a sample \CTMC\ path $\yV$ under a reversible, unnormalized rate matrix is:

\begin{eqnarray} 
\label{eq:aug_joint_den}
\log f_{\textrm{w}|\textrm{z,y}}(\wV|\zV,\yV) 
&\defeq& -\frac{1}{2} \kappa \Vert \wV \Vert^2_2  - \sum_{\x\in\statespace} \h_\x \sum_{\x'\in\statespace:\x\neq \x'} \qrev_{\x, \x'}\left(\wV\right)\nonumber,
 \\
  &&\;\;+\sum_{\x\in\statespace} \sum_{\x'\in\statespace:\x\neq \x'} \c_{\x, \x'} \log\left(\qrev_{\x, \x'}\left(\wV\right)\right) + 
\sum_{\x \in \statespace}n_x\log(\pi_{x}(\wV)).
\end{eqnarray}
In the $i$th iteration of  \LBPS-\HMC, while using \HMC\ kernel, we only update $\wV^u$ while conditioning on the rest. Thus, the exchangeable parameters are fixed at the values obtained in the $(i-1)$th iteration denoted as $\theta_{\{\x, \x'\}}\left(\left(\wV^b\right)^{i-1}\right)$. To simplify the notation, we denote $\theta_{\{\x, \x'\}}\left(\left(\wV^b\right)^{i-1}\right) $ as $\theta_{\{\x, \x'\}}$. We can  rewrite  Equation~\ref{eq:aug_joint_den} as:
\begin{eqnarray} 
\label{eq:uni_function_evaluation}
\log f_{\textrm{w}^{\textrm{u}}|\textrm{w}^\textrm{b},\textrm{z,y}}(\wV^{u}|\wV^b,\zV,\yV) 
&=& -\frac{1}{2} \kappa \Vert \wV^u \Vert^2_2  - \sum_{\x\in\statespace} \h_\x \sum_{\x'\in\statespace:\x\neq \x'} \pi_{x'}(\wV^u)\theta_{\{\x, \x'\}}
 \\
  &&\;\;+\sum_{\x\in\statespace} \sum_{\x'\in\statespace:\x\neq \x'} \c_{\x, \x'} \log\left(\pi_{x'}(\wV^u)\theta_{\{\x, \x'\}}\right) + 
\sum_{\x \in \statespace}n_x\log(\pi_{x}(\wV^u))\nonumber
\end{eqnarray}
Thus, the gradient of the augmented joint density with respect to $\wV^u$ is:
\begin{eqnarray} 
\label{eq:gradient_hmc_uni}
\nabla\log f_{\textrm{w}^{\textrm{u}}|\textrm{w}^\textrm{b},\textrm{z,y}}(\wV^{u}|\wV^b,\zV,\yV) 
&=& -\kappa \wV^u   - \sum_{\x\in\statespace} \sum_{\x'\in\statespace:\x\neq \x'} \h_\x 
\qrev_{\x, \x'}\left(\wV^u\right)\left(  \suffstatioV(x')-\sum_{\x\in\statespace} \pi_{x}(\wV^u) \suffstatioV(x)
\right)\nonumber
 \\
  &&\;\;+\sum_{\x\in\statespace} \sum_{\x'\in\statespace:\x\neq \x'} \c_{\x, \x'}\left( \suffstatioV(x')-\sum_{\x\in\statespace} \pi_{x}(\wV^u) \suffstatioV(x) \right) \nonumber\\
 &&\;\;  + 
\sum_{\x \in \statespace}n_x\left(\suffstatioV(x)-\sum_{\x\in\statespace} \pi_{x}(\wV^u)\suffstatioV(x)\right),
\end{eqnarray}
where:
\begin{eqnarray}
\nabla \A(\wV) &=& \sum_{\x\in\statespace} \suffstatioV(x) \pi_\x(\wV)\nonumber.
\end{eqnarray}

\section*{Appendix C. Gradient Computation for Univariate and Bivariate Weights using only \HMC\ kernel as benchmark sampling algorithm}
To investigate the computational efficiency of \LBPS-HMC, we choose state-of-the-art sampling algorithm \HMC\ to sample $\wV=\left(\wV^u, \wV^b\right)$. 
 
Since \HMC\ requires the gradient of $\wV$, we derive it in Equation~\ref{eq:aug_joint_den}:
\begin{eqnarray} 
\nabla\log f_{\textrm{w}|\textrm{z,y}}(\wV|\zV,\yV) 
&=& -\kappa \wV   - \sum_{\x\in\statespace} \sum_{\x'\in\statespace:\x\neq \x'} \h_\x 
\qrev_{\x, \x'}\left(\wV\right)\left(  \suffstatioV(x')-\sum_{\x\in\statespace} \pi_{x}(\wV) \suffstatioV(x)
\right)\nonumber
 \\
 &&\;\;-\sum_{\x\in\statespace} \sum_{\x'\in\statespace:\x\neq \x'} \h_\x\qrev_{\x, \x'}\left(\wV\right)\sufftransV(\{\x,\x'\})
 \nonumber\\ 
  &&\;\;+\sum_{\x\in\statespace} \sum_{\x'\in\statespace:\x\neq \x'} \c_{\x, \x'}\left( \suffstatioV(x')-\sum_{\x\in\statespace} \pi_{x}(\wV) \suffstatioV(x) \right) \nonumber\\
 &&\;\;+ \sum_{\x\in\statespace} \sum_{\x'\in\statespace:\x\neq \x'} \c_{\x, \x'}\sufftransV(\{\x,\x'\})
  + 
\sum_{\x \in \statespace}n_x\left(\suffstatioV(x)-\sum_{\x\in\statespace} \pi_{x}(\wV)\suffstatioV(x)\right)\nonumber.
\end{eqnarray}
\newpage
\section*{Appendix D. Software implementation correctness checks}

We outline in this section a simple extension of \cite{geweke2004getting}, which allows us to test the correctness of our software implementation. See \cite{geweke2004getting} for more background. We call the extension outlined in this section ``Exact Invariance Test'' (\EIT), which is described in more details in the website  \url{https://www.stat.ubc.ca/~bouchard/blang/Testing_Blang_models.html}.

As in \cite{geweke2004getting}, \EIT\ relies on comparing two sets of simulators, both targeting the joint distribution over the unobserved and observed variables. 
These methods exploit the fact that exact samples from this joint distribution can be obtained in two ways, described below. In what follows, we denote the prior by  $g_{\textrm{w}}$,  while $W'|W\sim K(\cdot|W)$ denotes the MCMC kernel. In our case, it is a Markov kernel based on HMC and LBPS. We also define a test function $t$.  	In our context, an example of $t(\cdot)$ is the function used to compute the exchangeable parameters given $W$. In general, $t$ could take as input both the unobserved and observed variables, but for simplicity it only depends on the unobserved one $W$ here. 

The first simulator consists in the following steps:
 
\begin{itemize}
\item For $m\in\{1, 2, \ldots, M_1\}$,
	\begin{itemize}
		\item Sample $W_m \sim g_{\textrm{w}}$. 
		\item Optional step: generate a dataset using the likelihood conditionally on $W_{m}$ (would only be needed if $t$ depends on both the synthetic data and $W$).
		\item $F_m=t(W_m)$. 	
	\end{itemize}
\end{itemize}

The second simulator consists in the following steps:

\begin{itemize}
  \item For $m\in\{1, 2, \ldots, M_2\}:$
	\begin{itemize}
		\item Sample $W_{1, m}\sim g_{\textrm{w}}$.
		\item Generate a dataset using the likelihood conditionally on $W_{1, m}$.
		\item For $j\in\{2, 3, \ldots, J\}$: $W_{j, m}|W_{j-1, m }\sim K(\cdot| W_{j-1, m}).$
		\item $H_m=t(W_{J, m}).$	
	\end{itemize}
\end{itemize}

For any $J\geqslant 1$, $a\in\{1, 2, \ldots, M_1\}$, $b\in\{1, 2, \ldots, M_2\}$, the random variable $F_a$ (generated from the first simulator) is equal in distribution to the random variable $H_b$ (generated by the second simulator) if and only if the kernel $K$ is $f$ invariant. This follows directly from the definition of global balance of Markov chains. This exact equality of distributions can then be used as the basis of a frequentist point-null hypothesis test.

\subsection*{\EIT\ Results}

Table~\ref{table:bps_eitt} and Table~\ref{table:bps_eit_exchangeable} show that our proposed \LBPS-HMC\  has passed the \EIT\ since all pvalues are bigger than a threshold of $0.05/n$, where $n$ represents the number of parameters in the test to take multiple comparisons into account. We show the test results of using \HMC\ alone in Table~\ref{table:hmc_eit} and Table~\ref{table:hmc_eit_exchangeable}. 

\begin{table}[H]
  \begin{center}
   \begin{tabular}{lccccc}
	 \toprule
	  &\multicolumn{5}{c}{Parameter Index}\\
	  \cmidrule(lrrrrr){2-6}
	   KS test& 1 & 2& 3&4&5\\
	   \midrule
	   Pvalue& 0.640&0.200&0.329&0.114&0.167\\
	   \midrule
	   Test statistics& 0.060&0.087&0.077&0.097&0.090\\
	   \bottomrule
  	\end{tabular}
   \end{center}
\caption{\EIT\ results for the weights of the stationary distribution using \LBPS-\HMC.}
   \label{table:bps_eitt}
\end{table}

\begin{table}[H]
	\begin{center}
	   \begin{tabular}{lcccccccccc}
	  \toprule
	  &\multicolumn{10}{c}{Parameter Index}\\
	  \cmidrule(lrrrrrrrrrr){2-11}
	KS test& 1 & 2& 3&4&5&6&7&8&9&10\\
	   \midrule
	   Pvalue & 0.200& 0.504 & 0.441&0.061 & 0.571&0.838 & 0.382&0.709&0.238& 0.382\\
	   \midrule
	   Test statistics& 0.087 & 0.067&0.070& 0.107 & 0.063&0.050&0.073&0.057&0.083&0.073\\
	   \bottomrule
	  \end{tabular}
   \end{center}
 \caption{\EIT\ results for exchangeable parameters using \LBPS-\HMC.}
 \label{table:bps_eit_exchangeable}
\end{table}


\begin{table}[H]
  \begin{center}
  	\begin{tabular}{lccccc}
	 \toprule
	  &\multicolumn{5}{c}{Parameter Index}\\
	  \cmidrule(lrrrrr){2-6}
	   KS test& 1 & 2& 3&4&5\\
	   \midrule
	   Pvalue& 0.967&0.200&0.640&0.838&0.640\\
	   \midrule
	   Test statistics& 0.040&0.087&0.060&0.050&0.060\\
	   \bottomrule
  	\end{tabular}
     \end{center}
 \caption{\EIT\ results for the weights of the stationary distribution  using \HMC\ alone.}
 \label{table:hmc_eit}
\end{table}

\begin{table}[H]
  \begin{center}
   \begin{tabular}{lcccccccccc}
	  \toprule
	  &\multicolumn{10}{c}{Parameter Index}\\
	  \cmidrule(lrrrrrrrrrr){2-11}
	  KS test& 1 & 2& 3&4&5&6&7&8&9&10\\
	   \midrule
	   Pvalue & 0.139& 0.076 & 0.504&0.281 & 0.838&0.640& 0.936&0.504&0.139& 0.238\\
	   \midrule
	   Test statistics& 0.093 & 0.103&0.067& 0.080 & 0.050&0.060&0.043&0.067&0.093&0.083\\
	   \bottomrule
	  \end{tabular}
   \end{center}
 \caption{\EIT\ results for exchangeable parameters using \HMC\ alone.}
 \label{table:hmc_eit_exchangeable}
\end{table}

\newpage
\section*{Appendix E. Posterior sample comparison between \LBPS-HMC\ and \HMC}

\begin{figure}[H]
  \begin{center}
   \includegraphics[scale=0.43]
    {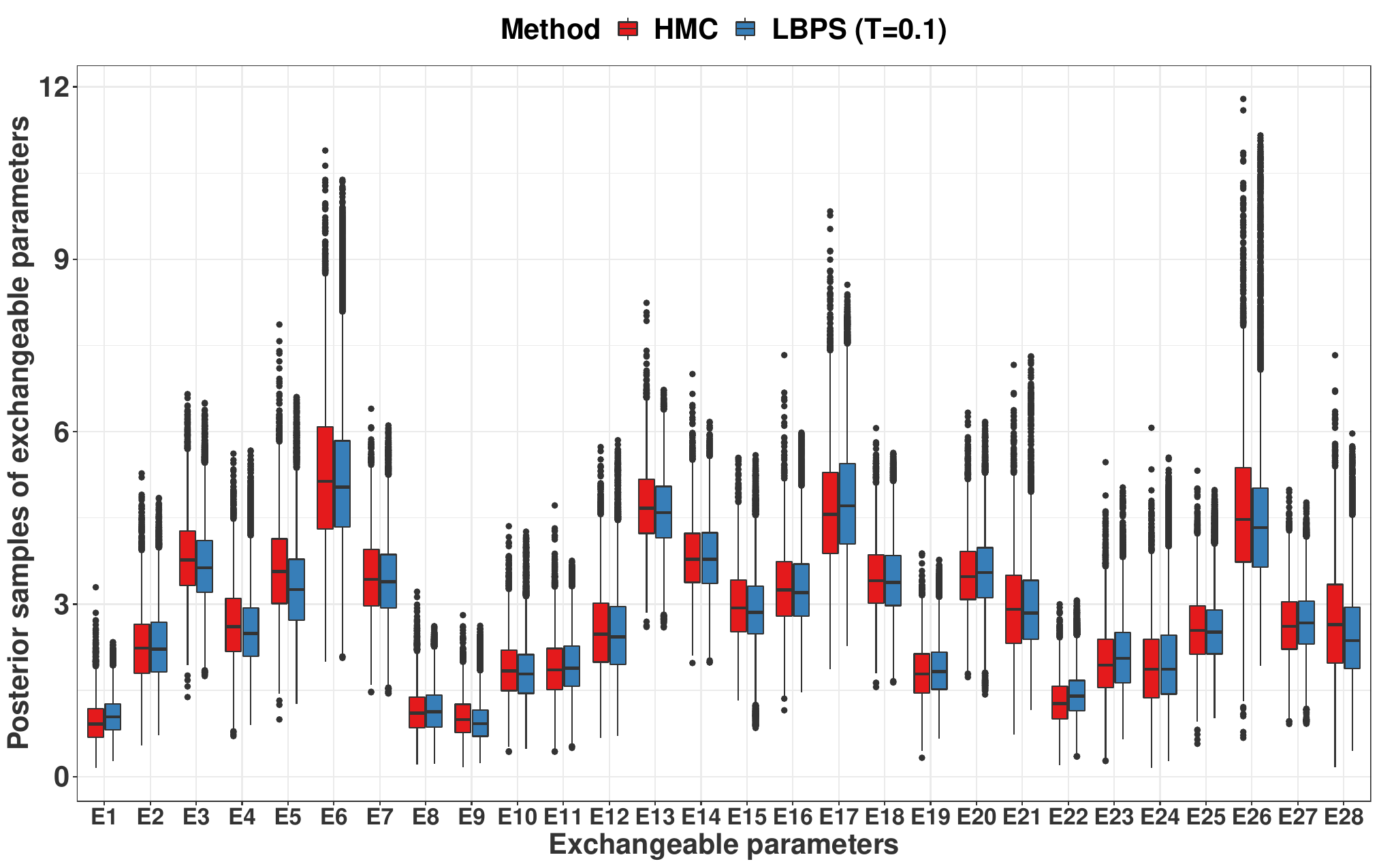}
  \end{center}
\caption{ Boxplot of posterior samples comparison for exchangeable parameters of an 8-by-8 rate matrix between \LBPS-HMC\ and \HMC.}
\label{fig:boxplot}
\end{figure}

In this section, we provide the summary statistics including the minimum, first quantile, median, mean, third quantile and maximum of the \ARD\,  defined in Equation~\ref{eq:ARD}, across all exchangeable parameters from one single run of \LBPS-\HMC\ and \HMC\ in Table~\ref{tb:abs_rela_diff} under the 8-by-8 Bayesian \GLM\ chain \GTR\ model described in Section~\ref{sec:model}. We also provide the boxplots of posterior samples collected from \LBPS-HMC\ and \HMC\ under the same set-up in Figure~\ref{fig:boxplot}.

\begin{table}[H]
  \begin{center}
   \begin{tabular}{cccccc}
   \toprule
       Min. & 1st Quantile& Median & Mean & 3rd Quantile & Max.\\
     \midrule
     1.4\%&1.16\%&2.70\%&3.05\%&3.63\%&12.06\%\\
    \bottomrule
    \end{tabular}
   \end{center}
 \caption{Summary of \ARD\ across exchangeable parameters between \HMC\ and \LBPS.}
 \label{tb:abs_rela_diff}
\end{table}

\newpage
\section*{Appendix F. Pairwise Euclidean Distance of Amino Acids}
\begin{center}
\begin{adjustbox}{angle=90}
\begin{tabular}{lrrrrrrrrrrrrrrrrrrrr}
	\hline
	  &   Y &   H &   Q &   R &   T &   N &   K &   D &   E &   G &   F &   L &   A &   S &   P &   I &   M &   V &   C &   W \\ \hline
	Y &   0 &  83 &  99 &  77 &  92 & 143 &  85 & 160 & 122 & 147 &  22 &  36 & 112 & 144 & 110 &  33 &  36 &  55 & 194 &  37 \\
	H &  83 &   0 &  24 &  29 &  47 &  68 &  32 &  81 &  40 &  98 & 100 &  99 &  86 &  89 &  77 &  94 &  87 &  84 & 174 & 115 \\
	Q &  99 &  24 &   0 &  43 &  42 &  46 &  53 &  61 &  29 &  87 & 116 & 113 &  91 &  68 &  76 & 109 & 101 &  96 & 154 & 130 \\
	R &  77 &  29 &  43 &   0 &  71 &  86 &  26 &  96 &  54 & 125 &  97 & 102 & 112 & 110 & 103 &  97 &  91 &  96 & 180 & 102 \\
	T &  92 &  47 &  42 &  71 &   0 &  65 &  78 &  85 &  65 &  59 & 103 &  92 &  58 &  58 &  38 &  89 &  81 &  69 & 149 & 128 \\
	N & 143 &  68 &  46 &  86 &  65 &   0 &  94 &  23 &  42 &  80 & 158 & 153 & 111 &  46 &  91 & 149 & 142 & 133 & 139 & 174 \\
	K &  85 &  32 &  53 &  26 &  78 &  94 &   0 & 101 &  56 & 127 & 102 & 107 & 106 & 121 & 103 & 102 &  95 &  97 & 202 & 110 \\
	D & 160 &  81 &  61 &  96 &  85 &  23 & 101 &   0 &  45 &  94 & 177 & 172 & 126 &  65 & 108 & 168 & 160 & 152 & 154 & 181 \\
	E & 122 &  40 &  29 &  54 &  65 &  42 &  56 &  45 &   0 &  98 & 140 & 138 & 107 &  80 &  93 & 134 & 126 & 121 & 170 & 152 \\
	G & 147 &  98 &  87 & 125 &  59 &  80 & 127 &  94 &  98 &   0 & 153 & 138 &  60 &  56 &  42 & 135 & 127 & 109 & 159 & 184 \\
	F &  22 & 100 & 116 &  97 & 103 & 158 & 102 & 177 & 140 & 153 &   0 &  22 & 113 & 155 & 114 &  21 &  28 &  50 & 205 &  40 \\
	L &  36 &  99 & 113 & 102 &  92 & 153 & 107 & 172 & 138 & 138 &  22 &   0 &  96 & 145 &  98 &   5 &  15 &  32 & 198 &  61 \\
	A & 112 &  86 &  91 & 112 &  58 & 111 & 106 & 126 & 107 &  60 & 113 &  96 &   0 &  99 &  27 &  94 &  84 &  64 & 195 & 148 \\
	S & 144 &  89 &  68 & 110 &  58 &  46 & 121 &  65 &  80 &  56 & 155 & 145 &  99 &   0 &  74 & 142 & 135 & 124 & 112 & 177 \\
	P & 110 &  77 &  76 & 103 &  38 &  91 & 103 & 108 &  93 &  42 & 114 &  98 &  27 &  74 &   0 &  95 &  87 &  68 & 169 & 147 \\
	I &  33 &  94 & 109 &  97 &  89 & 149 & 102 & 168 & 134 & 135 &  21 &   5 &  94 & 142 &  95 &   0 &  10 &  29 & 198 &  61 \\
	M &  36 &  87 & 101 &  91 &  81 & 142 &  95 & 160 & 126 & 127 &  28 &  15 &  84 & 135 &  87 &  10 &   0 &  21 & 196 &  67 \\
	V &  55 &  84 &  96 &  96 &  69 & 133 &  97 & 152 & 121 & 109 &  50 &  32 &  64 & 124 &  68 &  29 &  21 &   0 & 192 &  88 \\
	C & 194 & 174 & 154 & 180 & 149 & 139 & 202 & 154 & 170 & 159 & 205 & 198 & 195 & 112 & 169 & 198 & 196 & 192 &   0 & 215 \\
	W &  37 & 115 & 130 & 102 & 128 & 174 & 110 & 181 & 152 & 184 &  40 &  61 & 148 & 177 & 147 &  61 &  67 &  88 & 215 &   0 \\ \hline
\end{tabular}
\end{adjustbox}
\end{center}

\section*{Appendix G. Nearest Neighbour Pariwise Amino Acid Ordering }

\begin{algorithm}[H]
\caption{Nearest neighbour pairwise amino acid ordering}
\label{alg:ordering}
\begin{algorithmic}[1]
{
\small
\STATE \textbf{Initialization:} \\
$\statespace^{\textrm{dynamic, distinct}}= \statespace^{\textrm{distinct}}$, counter=0.\\AminoAcidPairRank=DynamicSupportForAminoAcid =dict(), which is an empty dictionary.\\
Set $\alpha$= $\beta$=$i_0$=$i_1$=NULL.
\FOR{$i\in\statespace$}
\STATE DynamicSupportForAminoAcid$[i]$ = $\statespace \setminus \{i\}$.
\ENDFOR
\WHILE{$\statespace^{\textrm{dynamic, distinct}}$ is not empty} 
\IF{$i_0$ is not NULL and $i_1$ is not NULL}
\STATE Find support of $i_0$: $\alpha$=DynamicSupportForAminoAcid$[i_0]$.
\STATE Find support of $i_1$: $\beta$=DynamicSupportForAminoAcid$[i_1]$.
\ENDIF
\IF{neither of $\alpha$ or $\beta$ is empty}
\STATE rowDist = AminoAcidDist[$i_0$, $\alpha$]
\STATE colDist = AminoAcidDist[$\beta$, $i_1$]
\STATE rowMinInd = $\alpha$[argmin(rowDist)]
\STATE colMinInd = $\beta$[argmin(colDist)]
	\IF{AminoAcidDist[$i_0$, rowMinInd] $\leqslant$ AminoAcidDist[colMinInd, $i_1$]}
		\STATE $i_1$ = rowMinInd
	\ELSIF{AminoAcidDist[$i_0$, rowMinInd] $>$ AminoAcidDist[colMinInd, $i_1$]}
		\STATE $i_0$ = colMinInd
	\ENDIF
\ELSIF{$\alpha$ is NULL and $\beta$ is not NULL}
	\STATE colDist=AminoAcidDist[$\beta$, $i_1$]
	\STATE colMinInd=$\beta$[argmin(colDist)]
	\STATE $i_0$= colMinInd
\ELSIF{$\alpha$ is not NULL and $\beta$ is NULL}
	\STATE rowDist = AminoAcidDist[$i_0$, $\alpha$]
	\STATE rowMinInd=$\alpha$[argmin(rowDist)]
	\STATE $i_1$ = rowMinInd
\ELSE
	\STATE  Find the amino acid pair SmallestPair $\in \statespace^{\textrm{dynamic,distinct}}$ with nonzero minimum in AminoAcidDist.
	\STATE $i_0$= index of the first amino acid in SmallestPair in $\statespace$.
	\STATE $i_1$ = index of the second amino acid in SmallestPair in $\statespace$.
\ENDIF
\STATE AminoAcidPairRank[$i_0$, $i_1$]=AminoAcidPairRank[$i_1$, $i_0$] =counter, counter ++.
\STATE AminoAcidDist[$i_0$, $i_1$]=AminoAcidDist[$i_1$, $i_0$]=$\infty$.
	\IF{$i_1$ $\in$ DynamicSupportForAminoAcid[$i_0$]}
		\STATE  Remove $i_1$ from DynamicSupportForAminoAcid[$i_0$]
	\ENDIF
	\IF{$i_0\in$ DynamicSupportForAminoAcid[$i_1$]}
		\STATE Remove $i_0$ from DynamicSupportForAminoAcid[$i_1$]
	\ENDIF
\STATE $s_0$ = $\statespace_{i_0} + \statespace_{i_1}$ (obtain amino acid pair $s_0$), $s_1$ = $\statespace_{i_1} + \statespace_{i_0}$ (obtain amino acid pair $s_1$).
	\IF{$s_0$ $\in$ $\statespace^{\textrm{dynamic, ordered,distinct}}$}
		\STATE Remove $s_0$ from $\statespace^{\textrm{dynamic, distinct}}$.
	\ENDIF
	\IF{$s_1$ $\in$ $\statespace^{\textrm{dynamic, distinct}}$}
		\STATE Remove $s_1$ from $\statespace^{\textrm{dynamic, distinct}}$
	\ENDIF
\ENDWHILE

\STATE Return AminoAcidPairRank.
}
\end{algorithmic}
\end{algorithm}

\section*{Appendix H. Traceplot comparison for real data analysis}\label{sec:violin}
We provide the traceplot on a randomly selected exchangeable parameter using \LBPS-HMC\ compared with \HMC. There is a small difference between posterior means with \ARD\ 0.43\%.
\begin{figure}[H]
\begin{center}
  \includegraphics[scale=0.5]
  {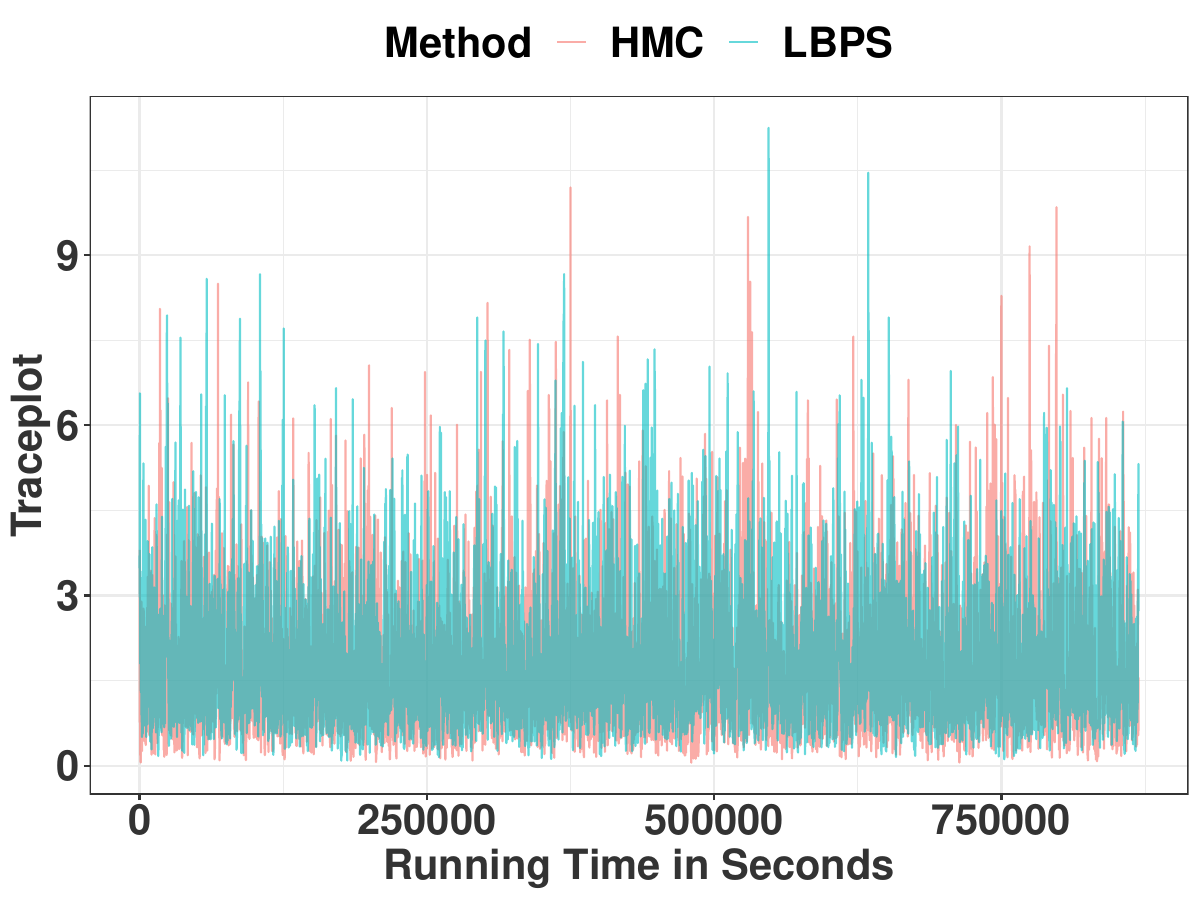}
 \end{center}
\caption{Traceplot for the 145th exchangeable parameter using \LBPS-\HMC\ compared to \HMC\ with \ARD\ 0.43\% in terms of the posterior mean.}
\label{fig:traceplot_real_data}
\end{figure}

\newpage
\section*{Appendix I. Correctness check via \ARD\ between two samplers}\label{sec:ard_difference}
With the same prior distribution and likelihood for the parameters (weights) of interest, \LBPS-\HMC\ and \HMC\ share the same posterior distribution. We evaluate the discrepancy among the posterior samples from the two algorithms using \ARD:

\begin{eqnarray}
\label{eq:ARD}
ARD(x, y)&=& \frac{\vert x-y\vert}{\textrm{max}(x, y)}, \, \textrm{for}\,\, \forall x, y>0.
 \end{eqnarray}

Figure~\ref{fig:ard_comp_real_data} and Table~\ref{tb:summary_ard} demonstrate the \ARD\ across all exchangeable parameters between \HMC\ and \LBPS-HMC\ with trajectory length 0.2. 

\begin{figure}[H]
  \begin{center}
   \includegraphics[scale=0.45]
   {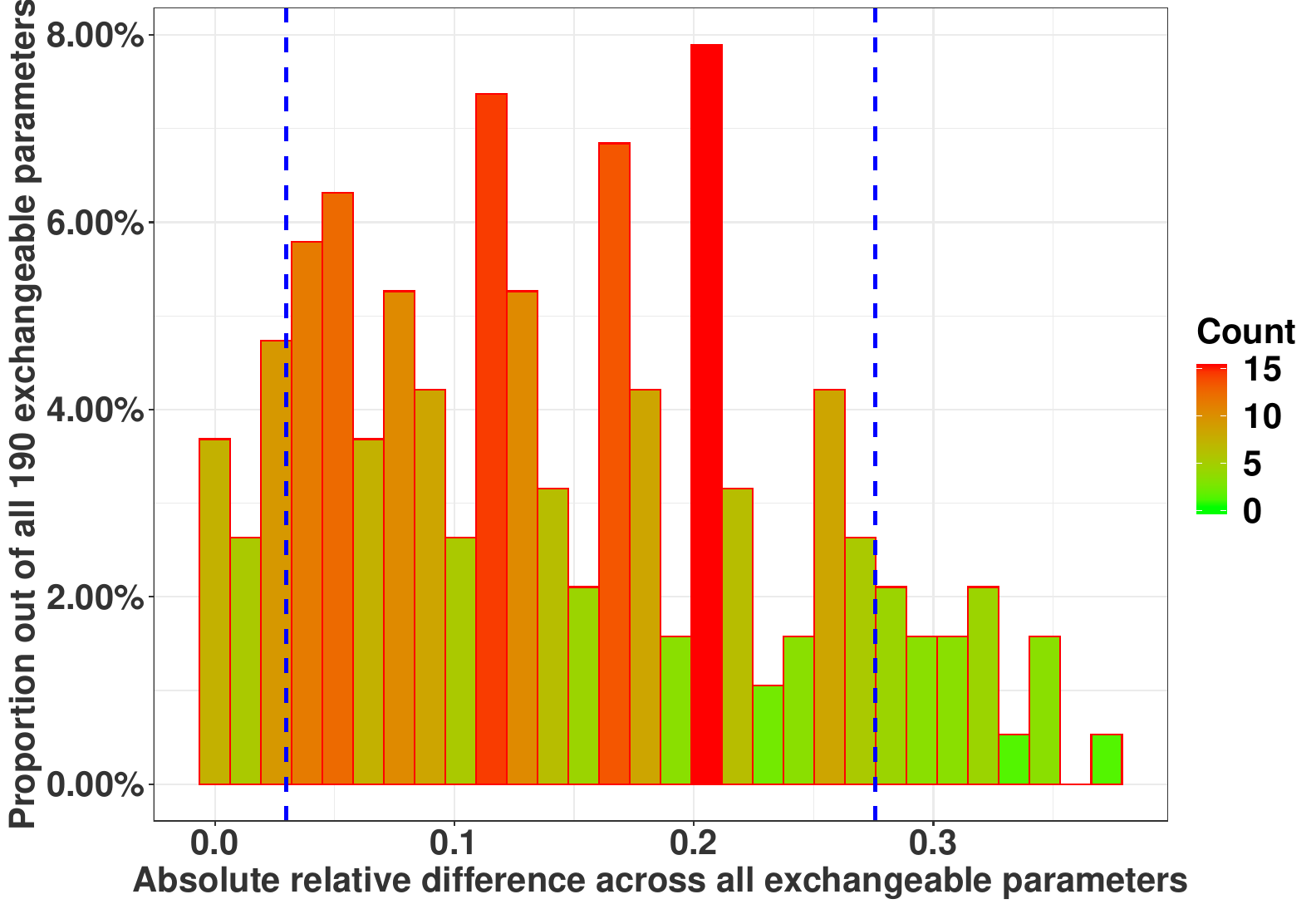}
  \end{center}
\caption{\ARD\ across exchangeable parameters between \LBPS-HMC\ compared to \HMC\ with blue dotted lines representing the 10\% and 90\% quantile of \ARD.}
\label{fig:ard_comp_real_data}
\end{figure}

\begin{table}[H]
  \begin{center}
   \begin{tabular}{lccccc}
   \toprule
     Min. & 1st Quantile& Median & Mean & 3rd Quantile & Max.\\
     \midrule
     0.3\%&   7.5\% &  14.8\%&   15.1\%  & 22.0\% &  44.4\%\\ 
     \bottomrule
    \end{tabular}
   \end{center}
 \caption{Summary of \ARD\ across exchangeable parameters.}
 \label{tb:summary_ard}
\end{table}

Since the maximum value of \ARD\ is 44.4\%, we further check whether it will decrease as expected if we run the chain longer. The discrepancy between the posterior samples from the two algorithms decreases if we have a longer chain. We validate this by providing a violin plot below comparing the \ARD\ from the current run denoted as ``shorter''with a longer run denoted as ``longer''. \HMC\ had a walltime of 12 days and \LBPS\ had a walltime of 10 days. Later, we ran each method for two more days. The maximum of \ARD\ dropped from 44\% to 37\% . 

We provide the quantile of \ESS\ in Table~\ref{table:ard_ess_quantile} and the actual \ESS\ in Table~\ref{table:ard_ess} of the exchangeable parameters that have \ARD\ bigger than its 95\% quantile between \LBPS-HMC\ and \HMC. We have found that a relatively bigger discrepancy between the posterior means occurs when either both methods have low \ESS\  or when \LBPS-HMC\ has high \ESS\ (such as 97.9\% quantile and 94.2\% quantile across all exchangeable parameters) but \HMC\ does not obtain an equivalently high \ESS\ for the same parameter.


\begin{figure}[H]
  \begin{center}
   \includegraphics[scale=0.2]
   {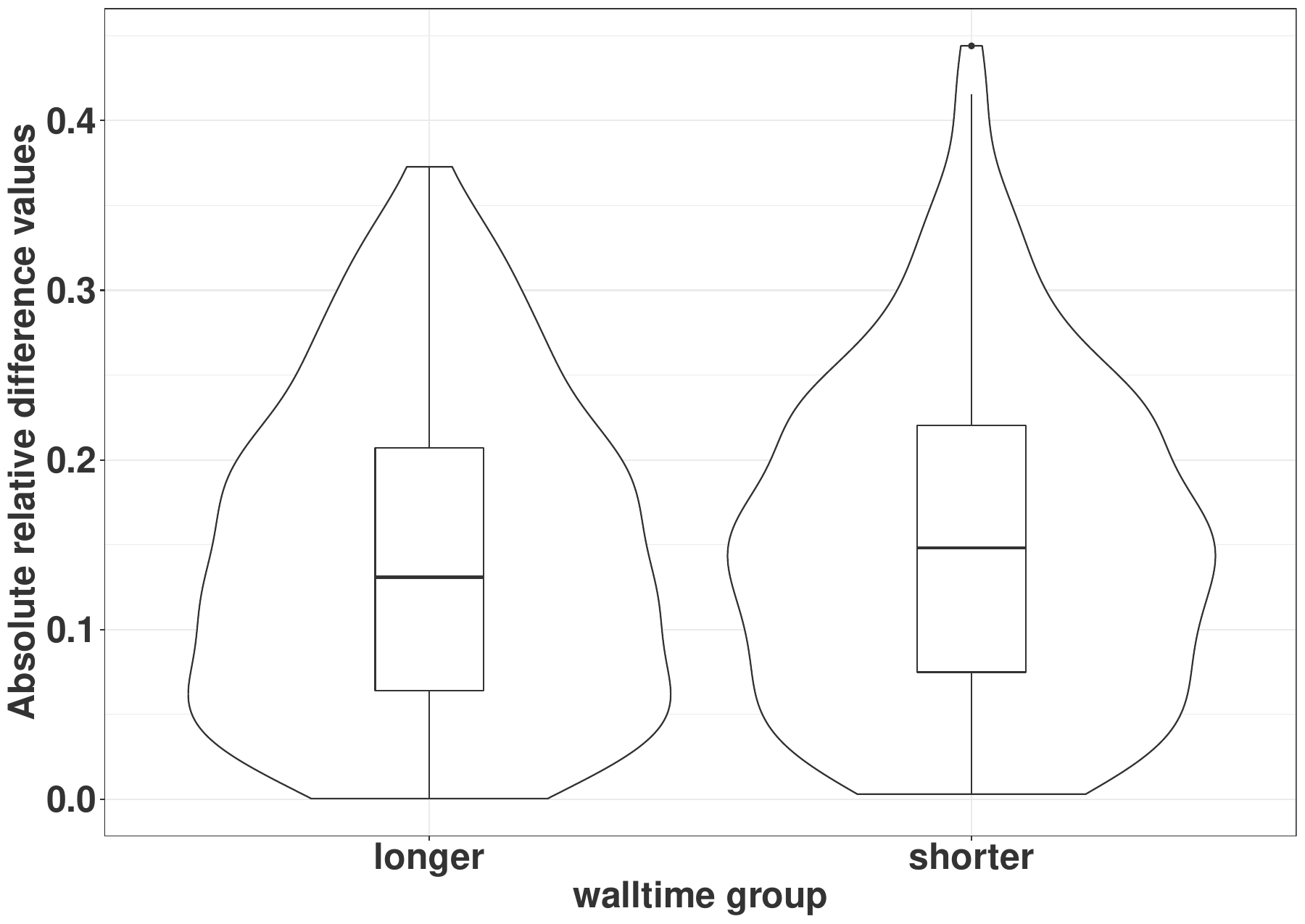}
  \end{center}
\caption{Violin plot of \ARD\ across exchangeable parameters between a ``shorter'' and ``longer'' run. }
\label{fig:ard_comp_real_data}
\end{figure}

\begin{table}[H]
  \begin{center}
   \begin{tabular}{lcccccccccc}
     \toprule
\LBPS-HMC\ \ & 2.63\%& 2.105\% &8.42\%& 5.79\%& 77.4\% & 94.2\%& 75.8\% &97.9 \% &0.526\% &3.16\%\\
  \midrule
  \HMC\ & 16.84\% &0.526\% &15.79\% &2.63\%& 53.7\%& 84.2\% &73.7\%& 98.4\%& 1.053\%& 5.26\%     \\
  \bottomrule
    \end{tabular}
   \end{center}
 \caption{Quantiles of \ESS\ of exchangeable parameters with \ARD\ bigger than its 95\% quantile.}
 \label{table:ard_ess_quantile}
\end{table}

\begin{table}[H]
  \begin{center}
   \begin{tabular}{lcccccccccc}
     \toprule
  \LBPS& 1071& 1051 &1248 &1147 &2145& 2330& 2115& 2451  &773 &1118\\
   \midrule
 \HMC& 544&  152&  604&  488 & 927& 1032& 1090&  591&  190 & 718\\ 
  \bottomrule
   \end{tabular}
   \end{center}
 \caption{The \ESS\ of exchangeable parameters with \ARD\ bigger than its 95\% quantile.}
 \label{table:ard_ess}
\end{table}

\newpage

\section*{Appendix J. Correctness of the sampling scheme LBPS-HMC}\label{sec:correctness_sampling_scheme}

In this section, we establish the correctness of our sampling scheme LBPS-HMC described in Algorithm~\ref{alg:algo_lbps}, i.e.\ that it is invariant with respect to the distribution with density $ g_{\textrm{w}|\textrm{y}}(\wV|\mathcal{Y}) $ in Equation~\ref{eq:posterior}. The sufficient statistics $Z$ defined in Section~\ref{sec:background} are used as ``transient auxiliary variable,'' i.e.\ quantities used to define a transition kernel but discarded after each transition is performed, a common technique in the MCMC literature, e.g.\ \citet{zhao2016bayesian}. 
We define $\boldsymbol{x}$ as the augmented sample path and a mapping $\boldsymbol{d}(\cdot)$ that takes $\boldsymbol{x}$ as input and outputs a vector containing the state of the \CTMC\ at each observation time, $\boldsymbol{y}$ as the states of the partially observed paths, and $g_{\mathrm{emi}}$ as an emission probability or density function. 

The joint distribution after auxiliary variable augmentation is 
\begin{align}
	g_{\mathrm{w}, \mathrm{x}, \mathrm{y}, \mathrm{z}}(\boldsymbol{w}, \boldsymbol{x}, \boldsymbol{y}, \boldsymbol{z}) &:= g_{\mathrm{w}}(\boldsymbol{w}) g_{\mathrm{x} \mid \mathrm{Q}}(\boldsymbol{x} \mid Q(\boldsymbol{w})) g_{\mathrm{y} \mid \mathrm{d}}(\boldsymbol{y} \mid \boldsymbol{d}(\boldsymbol{x})) \delta_{\boldsymbol{z}(\boldsymbol{x})}(\boldsymbol{z})
		\label{eq:joint} \\
g_{\mathrm{y} \mid \mathrm{d}}(\boldsymbol{y} \mid \boldsymbol{d}) &:= \prod_{\boldsymbol{v}\in \mathrm{V}} g_{\mathrm{emi}}\left(y_{v} \mid d_{v}\right).
\label{eq:emission}
\end{align}

Since $g_{\mathrm{x} \mid \mathrm{Q}}$ and $\delta_{\boldsymbol{z}(\boldsymbol{x})}$ integrate to one when marginalizing $\xV$ and $\zV$, we have that the augmented target admits the posterior distribution of interest ($g_{\textrm{w}|\textrm{y}}$, Equation~\ref{eq:posterior}) as a marginal. 

In the following, we define a Markov chain  $W^{(1)}, W^{(2)}, \ldots, $ on the state space $\mathbb{R}^P$. Our sampling scheme is a Markov transition kernel $T$ which is itself constructed using an alternation of several kernels $T_1, T_2, T_3$. We show that each  step in this sequence keeps $g_{\textrm{w}|\textrm{y}}(\wV|\mathcal{Y})$ invariant. It follows from standard results in MCMC theory that the alternation of these three kernels is also invariant with respect to $g_{\textrm{w}|\textrm{y}}$ \citep{tierney_markov_1994}.  A first kernel $T_1$ is used to instantiate the auxiliary variable $Z$. Then, a second kernel $T_2$ is used to perform changes in $W$ conditionally on $Z$, and finally, a last kernel $T_3$ projects the augmented space back to the target space $\mathbb{R}^P$. The broad lines of the arguments follow Appendix A in \citet{zhao2016bayesian}, with the key difference being that a combination of \LBPS\ and \HMC\ is used to design the kernel $T_2$ which updates $W$ given $Z$. We focus on the points in which the argument differs compared to \citet{zhao2016bayesian}.

Our sequence of transition kernels can be denoted as:
$\boldsymbol{w} \stackrel{T_{1}}{\longmapsto}(\boldsymbol{w}, \boldsymbol{z}) \stackrel{T_{2}}{\longmapsto}\left(\boldsymbol{w}^{\prime}, \boldsymbol{z}\right) \stackrel{T_{3}}{\longmapsto} \boldsymbol{w}^{\prime}$. Given the observations $Y$ and current weights $W$, the transition kernel $T_1$ performs the auxiliary variable $Z$ sampling with density $T_{1}\left(\boldsymbol{w}^{\prime}, \boldsymbol{z}^{\prime} \mid \boldsymbol{w}\right):= \delta_{\boldsymbol{w}}(\boldsymbol{w}^{\prime}) g_{\mathrm{z} \mid \mathrm{w}, \mathrm{y}}(\boldsymbol{z} \mid \boldsymbol{w}, \boldsymbol{y})$. Details of this sampling step are outlined in Section 4.3 and Supplement 3 of \citet{zhao2016bayesian}. 

The kernel $T_2$ performs several Metropolis-within-Gibbs steps on $\wV$ while keeping the auxiliary variable $\zV$ fixed. In \citet{zhao2016bayesian}, \AHMC\  is used to update all elements of $\wV$. In this paper, for reasons described in Section~\ref{sec:alternation}, we partition $\wV$ into two blocks: a first block corresponding to the univariate weights $\wV^{u}$ and a second block, to the bivariate weights $\wV^{b}$ (see Algorithm~\ref{alg:algo_lbps}). We use \HMC\ to update $\wV^u$ while fixing $\wV^{b}$ and $\zV$ and  \LBPS\ to update $\wV^b$ while fixing $\wV^{u}$ and $\zV$. This alternation can be denoted as $(\wV, \zV)=((\wV^u, \wV^b), \zV) \stackrel{\textrm{HMC}}{\longmapsto}({({\wV}^{\prime}}^u, \wV^b), \zV) \stackrel{\textrm{LBPS}}{\longmapsto} ({({\wV}^{\prime}}^u, {{\wV}^{\prime}}^b), \zV) =(\wV^{\prime}, \zV)$.
Again, as long as both the \HMC\ and \LBPS\ components are invariant with respect to an arbitrary conditional distribution of the augmented target $g_{\textrm{w}, \textrm{y}, \textrm{z}}$, their alternation is also guaranteed to preserve invariance. For \HMC, see for example \cite{Neal-10}, for \LBPS\ invariance was shown for an arbitrary distribution in \citet{bouchard2018bouncy}, Appendix F, so in particular it holds for a conditional distribution.

\vskip 0.2in

\bibliography{AllReferences}

\end{document}